\documentclass{llncs}

\usepackage{url}
\usepackage{paralist}
\usepackage{multicol}
\usepackage{multirow}
\usepackage{algorithm}
\usepackage{algorithmic}
\usepackage{graphicx}
\usepackage{pstricks}
\usepackage{epsfig}
\usepackage{amsmath}
\usepackage{amssymb}
\newcommand{\acronym}{CoCoE}
\newcommand{\ie}{\textit{i.e.},~}
\newcommand{\eg}{\textit{e.g.},~}
\newcommand{\cf}{\textit{c.f.},~}
\newcommand{\ua}{$\uparrow$}
\newcommand{\da}{$\downarrow$}

\makeatletter
\def\setliststart#1{\setcounter{\@listctr}{#1}%
  \addtocounter{\@listctr}{-1}}
\makeatother

\begin{document}

\title{\acronym: A Methodology for Empirical Analysis of LOD 
Datasets\thanks{This work has been supported by the `KI2NA' 
project funded by Fujitsu Laboratories Limited in collaboration with Insight @ 
NUI Galway. We also want to thank V\'aclav Bel\'ak for valuable discussions
regarding applicable clustering algorithms.}}

\author{V\'it Nov\'a\v{c}ek}
\institute{
Insight @ NUI Galway (formerly known as DERI)\\
IDA Business Park, Lower Dangan, Galway, Ireland\\
E-mail: \url{vit.novacek@deri.org}\\}

\maketitle

\begin{abstract}
\acronym~stands for Complexity, Coherence and Entropy, and presents an
extensible methodology for empirical analysis of Linked Open Data (\ie
RDF graphs). \acronym~can offer answers to questions like: {\it Is 
dataset A better than B for knowledge discovery since it is more complex and 
informative?}, {\it Is dataset X better than Y for simple value lookups due 
its flatter structure?}, etc. In order to address such questions, we introduce 
a set of well-founded measures based on complementary notions from 
distributional semantics, network analysis and information theory. These 
measures are part of a specific implementation of the \acronym~methodology 
that is available for download. Last but not least, we illustrate \acronym~by
its application to selected biomedical RDF datasets.
\end{abstract}

\section{Introduction}\label{sec:introduction}

As the LOD cloud is growing, people increasingly often face the problem of 
choosing the right dataset(s) for their purposes. Data publishers usually 
provide descriptions that can indicate possible uses of their datasets, 
however, such asserted descriptions may often be too shallow, subjective or
vague. 
To complement the dataset descriptions authored by their creators, maintainers 
or users, we introduce a comprehensive set of empirical quantitative measures 
that are based on the actual content of the datasets. Our main goal is to 
provide means for comparison of RDF datasets along several well-founded 
criteria, and thus determine the most appropriate datasets to utilise in 
specific use cases.

To motivate and illustrate our contribution from a practical point of view,
imagine a researcher Rob working on a novel method for discovering drug side 
effects. Rob knows that the most successful methods typically define and train 
a model in order to discover unknown side effects of drugs using their known 
features~\cite{harpaz2012novel}. Rob also knows there are datasets in the LOD 
cloud that can be used for defining features that may not be captured by the 
state of the art approaches. Moreover, due to the common RDF format, the 
datasets can relatively easily be combined to generate completely new sets of 
features. Therefore using relevant LOD data can lead Rob to 
a breakthrough in 
adverse drug effect discovery. 

Examples of such 
data 
are DrugBank, SIDER and Diseasome (\cf 
\url{http://datahub.io/dataset/fu-berlin-[drugbank|sider|diseasome]}). 
They des\-c\-ribe drugs, medical conditions, 
genes, etc. 
The question is 
how to use the datasets efficiently. 
Rob may 
wonder how much information can he typically gain from the datasets and which 
one is the best in this respect. Which 
of them is better for extracting flat features based on predicate-object 
pairs, and which is better for features based on more complex structural 
patterns? 
Last but not least, it may be useful to know what happens if one combines the 
datasets. Maybe it will bring more interesting features, and maybe
nothing will change much, only the data will become larger and more difficult 
to process.

\acronym~provides a well-founded methodology for empirical analysis of RDF 
data which can be used to determine applicability of the data to particular 
use cases (like Rob's in the motivating example above). The methodology is 
based on sampling the datasets with quasi-random 
heuristic walks that simulate various exploratory strategies of real agents. 
For each sample (\ie walk), we compute a set of measures that are then 
averaged across all the samples to approximate the overall characteristics 
of the dataset. We define three types of measures: complexity, coherence and
entropy. 
The purpose of the measures is to assess datasets along 
complementary 
perspectives that can be quantified in a well-founded and easy-to-interpret
manner. The perspectives chosen and their possible combinations cover a broad
area of use cases in which RDF datasets can possibly be applied, ranging from
simple value look-ups through semantic annotations to complex knowledge 
discovery tasks. 

The \acronym~methodology 
can obviously be implemented in many different ways, but here 
we describe only one specific realisation. For the complexity measures, we 
use network analysis algorithms~\cite{sna}. 
For the coherence measures, two auxiliary structures are required. Firstly, we 
need a distributional representation of the RDF data~\cite{Novacek2011iswc},
which describes each entity (subject or object) by a vector that represents 
its meaning based on the entities linked to it. Secondly, we need a taxonomy 
of nodes in 
the RDF (multi)graph, which is computed from the data itself by means of 
nonparametric hierarchical clustering. These two structures allow for 
representing coherence using various types of semantic similarities based on 
the vector space representation and the taxonomy structure, such as cosine or 
Wu-Palmer~\cite{10.1371/journal.pcbi.1000443}. The taxonomy structure also
serves as a basis for the 
entropies 
computed using cluster
annotations of the nodes in the walks. 

The rest of the paper is organised as follows. Section~\ref{sec:related} 
summarises the related work. Details on the \acronym~methodology and its 
implementation are given in Section~\ref{sec:methods}. 
Section~\ref{sec:experiments} presents an experimental illustration of the
\acronym~approach. 
We conclude the paper 
in Section~\ref{sec:conclusions}.


\section{Related Work}\label{sec:related}

The distributional representation of RDF data we use builds on our 
previous work~\cite{Novacek2011iswc}. We have recently introduced the notion 
of heuristic quasi-random walks and their empirical analysis 
in~\cite{novacek2014peerj}, which, however, deals with different types of
data, manually curated taxonomies and predefined gold standards. The presented
paper extends that work into a generally applicable methodology for analysing
RDF datasets using only the data itself. 

The clustering method introduced here builds on principles similar to k-hop 
clustering~\cite{nocetti2003connectivity}. Another related approach is 
nonparametric hierarchical link clustering~\cite{ahn2010link}, which is more
general and sophisticated than our simple method, yet its Python implementation
we have experimented with proved to be intractable when used in our
experiments. A comprehensive overview of semantic similarity measures 
applicable in \acronym~is provided in~\cite{10.1371/journal.pcbi.1000443}. The 
similarities used in our experiments were the cosine and Wu-Palmer ones, 
chosen as representatives of the vector space-based and taxonomy-based 
similarity types.

The most relevant tools and approaches for RDF data analysis 
are~\cite{campinas2012introducing,DBLP:conf/dexaw/LangeggerW09,moller2010learning,colazzo:hal-00960609}. 
Perhaps closest
to our work is~\cite{DBLP:conf/dexaw/LangeggerW09} that computes a set of
statistics and histograms for a given RDF dataset. The statistics are, 
however, concerned mostly about distributions of statements, instances and 
explicit statement patterns. This may be useful for tasks like SPARQL query 
optimisation, but 
cannot directly answer questions that motivate our work. Graph 
summaries~\cite{campinas2012introducing} propose high-level abstractions of 
RDF data intended to facilitate formulation of SPARQL queries, which is 
orthogonal to our approach aimed at quantitative characteristics of the data
itself. 
Usage-based RDF data analysis~\cite{moller2010learning} provides insights into 
common patterns of utilising RDF data by agents, but does not offer means for 
actually analysing the data. 
Finally, the recent 
approach~\cite{colazzo:hal-00960609} is useful for knowledge discovery in 
RDF data based on user-defined query patterns and analytical perspectives. 
Our approach complements~\cite{colazzo:hal-00960609} by characterising 
application-independent features of RDF datasets taken as a whole.


\section{Methods}\label{sec:methods}


In this section, we first introduce 
various RDF data representations that underlie the \acronym~methodology. Then 
we describe the 
clustering method used for computing taxonomies that are needed for certain
\acronym~measures. The concept of heuristic quasi-random walks is described 
then, followed by details on the \acronym~measures. Finally, we explain how
to interpret the measure values. 

\vspace{-0.3cm}
\subsection{Representations of RDF Datasets}\label{sec:methods.representations}
\vspace{-0.1cm}

\begin{figure}[ht]
\center
\scalebox{0.2}{\includegraphics{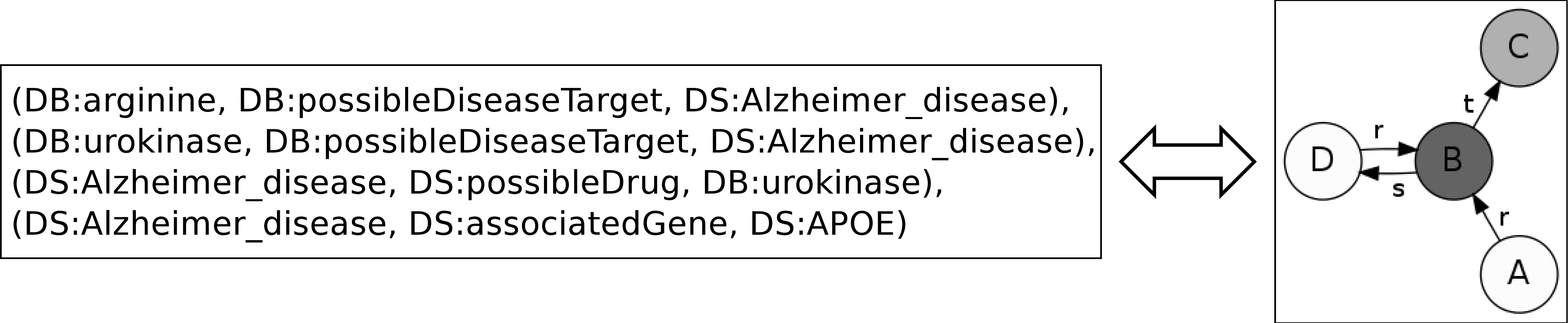}}
\caption{Example of a direct graph representation of RDF data}\label{fig:graph}
\end{figure}

Let us assume an RDF dataset consisting of triples $(s,p,o)$ that range over a 
set of URIs $\mathcal{U}$ and literals $\mathcal{L}$ such that $s, p \in 
\mathcal{U}, o \in \mathcal{U} \cup \mathcal{L}$. 
A {\bf direct graph representation} of the dataset is a directed labelled 
multigraph 
$G_d = (V,E_d,L_d)$ where $V = \mathcal{U} \cup \mathcal{L}$ is a set of 
nodes corresponding to the subjects and objects, $E_d$ is a set of ordered 
pairs $(u,v) \in V \times V$, and $L_d: E_d \rightarrow \mathcal{U}$ is a 
function that assigns a predicate label $p$ to every edge $(s,o)$ such that 
$(s,p,o)$ exist in the dataset. Note that we do not distinguish between URI 
and literal objects in the current implementation of \acronym~as we are 
interested in the most generic schema-independent features of the datasets. 
An example of a direct graph representation is given in Figure~\ref{fig:graph}. 
The left hand side of the figure contains RDF statements coming from DrugBank 
and Diseasome. For better readability, we represent the statements as simple
tuples, with the DB and DS abbreviations referring to the corresponding 
namespaces. We also 
use symbolic names instead of alphanumeric IDs present in the original data. 
In the graph representation in the right hand side of the figure, 
{\tt arginine}, {\tt Alzheimer\_disease}, {\tt APOE} and {\tt urokinase} 
correspond to the {\tt A}, {\tt B}, {\tt C} and {\tt D} codes, respectively. 
Similarly, the predicates {\tt possibleDi\-sea\-seTarget}, {\tt possibleDrug} 
and {\tt associatedGene} correspond to the {\tt r}, {\tt s} and {\tt t} codes, 
respectively. Drug entities are displayed in white, while disease and gene are 
in dark and light grey, respectively. 

Next we define a {\bf distributional representation} of an RDF dataset 
as a matrix $M$. The row indices of $M$ correspond to the set $V$ of nodes 
in the dataset's direct graph representation $G_d$. The column indices 
represent the context of the nodes by means of their connections in the $G_d$ 
graph, and are defined as a union of two sets of pairs corresponding to 
all possible outgoing and incoming edges: $\{(L_d((x,y)),y)| x \in V 
\wedge (x,y) \in E_d\} \cup \{(x,L_d((x,y)))| y \in V \wedge (x,y) \in 
E_d\}$. The values of the $M$ matrix indexed by a row $a$ and column $(b,c)$ 
are $1$ if there is an edge $(a,c)$ with predicate label $b$ or an edge $(b,a)$
with predicate label $c$ in $G_d$, and $0$ otherwise.
The dataset from the previous example corresponds to the following 
distributional representation:
\begin{center}
{\tt \scriptsize
\begin{tabular}{l|ccccccc}
  & (r,B) & (s,D) & (t,C) & (A,r) & (B,s) & (B,t) & (D,r) \\
\hline
A &   1   &   0   &   0   &   0   &   0   &   0   &   0   \\
B &   0   &   1   &   1   &   1   &   0   &   0   &   1   \\
C &   0   &   0   &   0   &   0   &   0   &   1   &   0   \\
D &   1   &   0   &   0   &   0   &   1   &   0   &   0   \\
\end{tabular}.}
\end{center}
The rows of the matrix can be used for computing similarities between 
particular data items using measures like cosine distance. Let us use a 
notation $\vec{x}$ to refer to a row vector in $M$ corresponding to the 
entity $x$ (\ie a subject or object in the original dataset). Then the 
cosine similarity between two entities $x,y$ is:
$
sim_{cos}(x,y) = \frac{\vec{x}\cdot\vec{y}}{|\vec{x}||\vec{y}|}.
$
For instance, the similarity between {\tt A, D} (\ie arginine and urokinase) is
$sim_{cos}(\mathtt{A},\mathtt{D}) = \frac{1}{\sqrt{1}\sqrt{2}} \simeq 0.707$.

For larger datasets, it is practical to use {\bf dimensionality reduction} to 
facilitate computations utilising the distributional representation. In the 
experiments presented here, we used a simple method for ranking and filtering
out the columns -- the $\chi^2$ statistic~\cite{statistics_for_research}, 
which can be used for computing divergence of specific observations from 
expected values. In our case, observations are the columns in a distributional 
representation, and their values are the column frequencies in the data. More 
formally, let us assume an $m \times n$ distributional representation $M$ with 
row index set $V = \{v_1,v_2,\dots,v_m\}$ and column index set $C = 
\{c_1,c_2,\dots,c_n\}$. Then the expected (\ie mean) and observed values for 
the $\chi^2$ statistic are $E(M) = \frac{1}{|C|}\sum_{r \in V, c \in C} 
M_{r,c}$, and $O(c_i,M) = \sum_{r \in V} M_{r,c_i}$ (for a column $c_i$). 
Using these formulae, the $\chi^2$ statistic of a column $c_i$ is 
$\chi^2(c_i,M) = \frac{(O(c_i,M) - E(M))^2}{E(M)}$.
The $\chi^2$ values for the columns in our example distributional 
representation are as follows. The expected value is $\frac{8}{7} \simeq 1.14$
(sum of all values in the matrix divided by the number of columns). All the 
columns but {\tt (r,B)} have $\chi^2$ value of $\frac{1}{56} \simeq 0.02$. The 
{\tt (r,B)} column has $\chi^2$ value of $\frac{9}{14} \simeq 0.64$. Therefore
one can consider {\tt (r,B)} as the only significant column. The similarity 
between the {\tt A} and {\tt D} entities then increases to $1$ as their 
corresponding vectors, reduced to the only significant dimension, are equal. 

\begin{figure}[ht]
\center
\scalebox{0.25}{\includegraphics{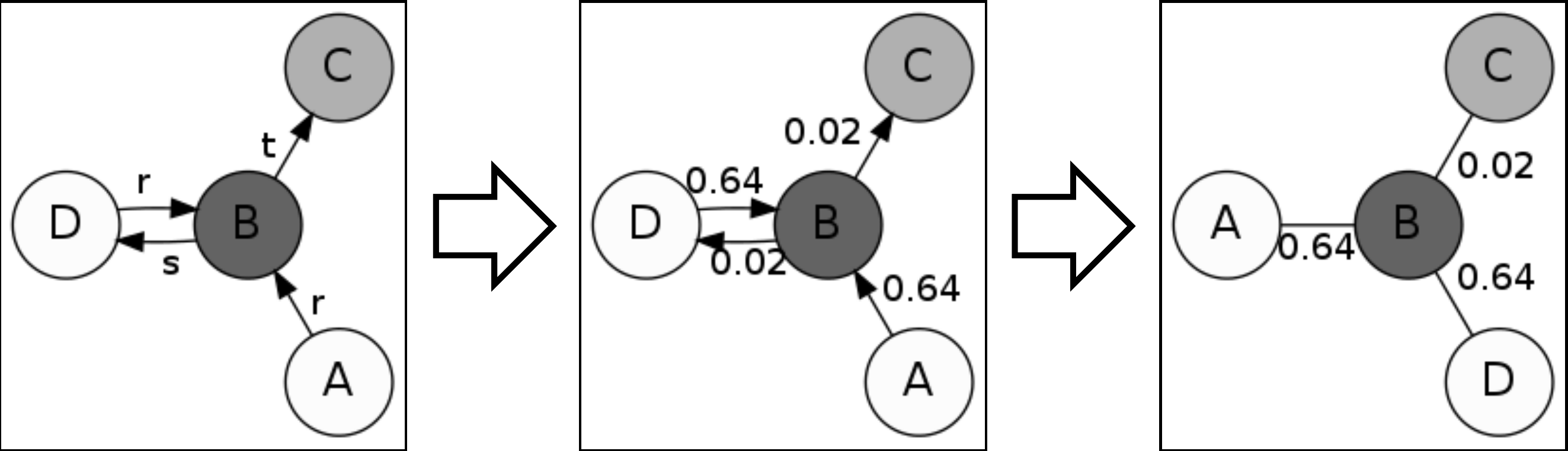}}
\caption{Example of a weighted indirect graph representation of RDF 
data}\label{fig:wgraph}
\end{figure}

In addition to reducing the dimensionality, 
we use the $\chi^2$ scores to construct {\bf weighted 
indirect representations} of RDF datasets, $G_w = (V,E_w,L_w)$. $G_w$ is an 
undirected graph with node set $V$ and edge set $E_w$ that consists of 
2-multisets of elements from $V$. 
The $E_w$ set is constructed from the 
corresponding direct graph representation $G_d$ as $\{\{u,v\} | (u,v) \in E_d 
\vee (v,u) \in E_d\}$. The labeling function $L^w_u: E_w \rightarrow 
\mathbb{R}$ associates the edges with a weight that is computed as maximum
from the values 
$\{\chi^2((p,u),M)|p \in P_I\} \cup 
 \{\chi^2((p,v),M)|p \in P_O\} \cup
 \{\chi^2((u,p),M)|p \in P_O\} \cup
 \{\chi^2((v,p),M)|p \in P_I\}$, where $P_I, P_O$ are sets of RDF predicates 
linking $v$ to $u$ and $u$ to $v$, respectively.
Figure~\ref{fig:wgraph} shows how the direct graph representation can be turned
into the indirect weighted one. 
\vspace{-0.3cm}
\vspace{-0.3cm}

The last structure we need for computing the \acronym~measures is a {\bf 
similarity representation} $G_s = (V,E_s,L_s)$. 
Similarly to $G_w$, $G_s$ is a weighted undirected graph. It captures
the similarities between the entities in the corresponding RDF dataset. The
edge set $E_s$ is defined as $\{\{u,v\} | sim_{cos}(\vec{u},\vec{v}) > 
\epsilon\}$, where $\vec{u},\vec{v}$ are the vectors in the dataset's 
distributional representation $M$ and $\epsilon \in [0,1)$ is a threshold. The 
edge labeling function $L_s: E_s \rightarrow (0,1]$ then assigns the actual 
similarities to the particular edges. Our example dataset has a sparse 
similarity representation, as most of the similarities are $0$, except of 
$sim_{cos}(\vec{A},\vec{D}) = sim_{cos}(\vec{D},\vec{A}) = 1$ (or 
$\frac{\sqrt{2}}{2}$ when using all dimensions).

\subsection{Nonparametric Hierarchical Clustering}\label{sec:methods.clustering}

To compute many of the \acronym~measures, a taxonomy of the nodes in the RDF
data representations is required. In some domains, standard, manually curated 
taxonomies exist (such as MeSH in life sciences, \cf 
\url{http://download.bio2rdf.org/current/mesh/mesh.html}). Unfortunately, such 
authoritative resources are not available for most domains, or they do not 
cover many RDF datasets sufficiently. Therefore we devised a simple 
algorithm that computes a hierarchical cluster structure (\ie taxonomy) based
on traversing the graph representations of the data. We compute two 
taxonomies $T_w, T_s$ based on the $G_w, G_s$ representations, respectively. 
$T_w$ is based on the data representation directly,
while $T_s$ captures 
the taxonomy induced by the entity similarities.

The most specific (\ie leaf-level) clusters are computed as follows (using the 
corresponding $G_? = (V,E_?,L_?)$ representation where $?$ is one of $w,s$):
\begin{enumerate}
  \item Compute a list $L$ of nodes $v \in V$ ranked according to their 
  clustering coefficients 
  $\frac{2\lambda_{G_?}(v)}{|a(v)|(|a(v)|-1)}$, 
  where $\lambda_{G_?}(v)$ is the number of complete subgraphs of $G_?$ 
  containing $v$, and $a(v)$ is a set of neighbours of $v$ in $G_?$ (we use 
  clustering coefficient as a simple quantification of node complexity and 
  their potential for spawning clusters).
  \item Set a cluster identifier $i$ to $0$ and initialise a mapping $\nu: 
  \mathbb{N} \rightarrow 2^V$ between cluster identifiers and corresponding 
  node sets. 
  \item While $L$ is not empty, do:
  \begin{enumerate}
    \item Pick a node $x$ with the highest rank from $L$.
    \item Set cluster $\nu(i)$ to a set of nodes $\{u | \Pi_{e \in p(x,u)} 
    L_?(e) > \epsilon\}$ where $p(x,u)$ is a set of edges on a 
    path between the nodes $x,u$ in $G_?$, and 
    $\epsilon$ is a predefined threshold (in our experiments, we set the 
    $\epsilon$ threshold dynamically to ca. $75$th percentile of the actual 
    edge weights in the given graph).
    \item Remove all $\nu(i)$ nodes from $L$ and increment $i$ by $1$.
  \end{enumerate}
  \item Return the cluster-to-nodes mapping $\nu$.
\end{enumerate}
Clusters of level $k$ are computed using the above algorithm from clusters at
the level $k-1$, continually adding new cluster identifiers. The algorithm is, 
however, applied on an undirected weighted cluster graph $G_c = (V_c,E_c,L_c)$
and generates the higher-level clusters by unions of nodes associated with the 
lower level ones. The $G_c$ graph for a level $k$ is 
defined as follows. Let us assume that the level $k-1$ consists of $n$ clusters
$c_1, c_2, \dots, c_n$ that correspond to sets of nodes $\nu(c_1), \nu(c_2), 
\dots, \nu(c_n)$. Then the node set $V_c$ for level $k$ equals to 
$\{c_1, c_2, \dots, c_n\}$ and the edge set $E_c$ is computed as $\{\{u,v\} | 
\exists x \exists y . x,y \in V_c \wedge (u \in \nu(x) \cap \nu(y) \vee v \in 
\nu(x) \cap \nu(y))\}$. The weight labeling $L_c$ assigns a weight to each 
edge in $E_c$ according to the following formula: $L_c(\{x,y\}) = 
\frac{1}{3}(2\sum_{e \in E^*} L_?(e) + \sum_{e \in E^+} L_?(e))$, where $L_?$ 
is the weight labeling function of the corresponding $G_?$ graph 
representation, and $E^*, E^+$ are sets of edges in $G_?$ that are fully and 
partially covered by the nodes in the $\nu(x) \cap \nu(y)$ intersection (full 
coverage means that both nodes of an edge are in the intersection, while the 
partial coverage requires exactly one edge node to be present there). It is
easy to see that the $G_c$ graphs connect clusters that have non-empty node 
overlap. The weights of the connections are computed as a weighted 
arithmetic mean of the weights of edges with nodes in the cluster 
intersections, where the edges with both nodes in the intersection contribute
twice as much as the edges with only one node there. 

The final product of the clustering algorithm is a mapping between cluster
identifiers and corresponding sets of nodes. As the more specific (\ie 
lower-level) clusters are incrementally merged into more abstract ones, each 
node can be assigned a set of so called tree codes that reflect its membership 
in the particular clusters. The tree codes have the form 
$L_1.L_2.\;\dots\;.L_{n-1}.L_n$ where $L_i$ are identifiers of clusters of
increasing specificity (\ie $L_1, L_n$ are the most general and specific,
respectively). For the \acronym~measures, we sometimes consider only the 
top-level cluster identifiers which we denote by $C^X_T$ for an entity 
{\tt X}. The notation $C^X_S$ refers to the set of all specific cluster 
identifiers associated with an entity {\tt X}. 

\begin{figure}[ht]
\center
\scalebox{0.25}{\includegraphics{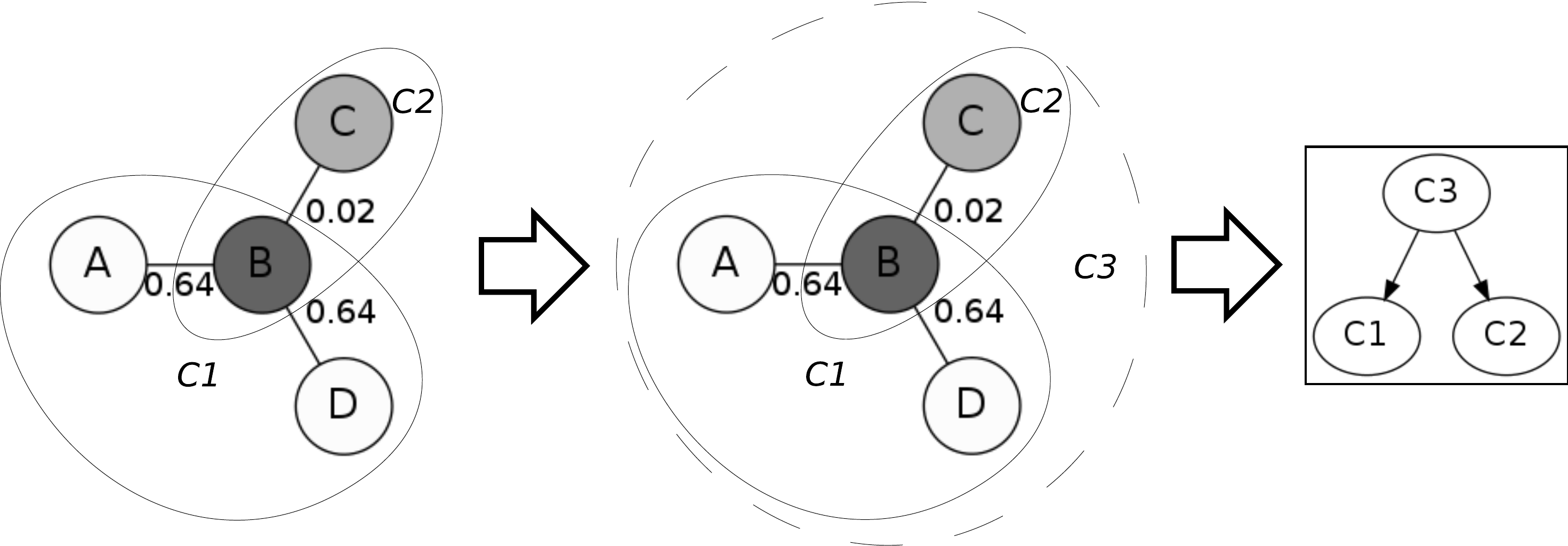}}
\caption{Example of a clustering}\label{fig:clustering}
\end{figure}

To give an example of how the clustering works, let us assume the $\epsilon$ 
threshold is set to the minimum of the graph weights at each level of the 
clustering. Considering the dataset from the previous examples, the 
computation of the initial clusters according to the $G_w$ representation 
can start from any node as their clustering coefficient is always zero. Let us 
start from the node {\tt A} then. The corresponding hierarchical clustering 
process is depicted in Figure~\ref{fig:clustering}, together with the 
resulting cluster structure (\ie dendrogram).
\vspace{-0.3cm}
\vspace{-0.3cm}
The value of $\epsilon$ in the first step is $0.02$ and thus the 
clustering puts the nodes {\tt A, B, D} into a cluster {\tt C1} first, 
proceeding from {\tt C} then and creating a cluster {\tt C2} consisting of
{\tt C, B}. No other traversals are possible as the multiplied edge weights 
fall below the threshold already. The next level uses the \{{\tt B,C}\} edge 
weight $0.02$ as the $\epsilon$ threshold again as it is the only edge 
connecting the {\tt C1, C2} clusters. The top-most cluster {\tt C3} is a union 
of the {\tt C1, C2} ones. The resulting sets of tree codes are: 
\{{\tt C3.C1}\} for nodes {\tt A, D}; \{{\tt C3.C2}\} for 
node {\tt C}; \{{\tt C3.C1, C3.C2}\} for node {\tt B}. 

For some of the measures defined later on, we need a notion of the {\bf number 
and size of clusters}. Let us assume a set of entities $Z \subseteq V$. The 
number of clusters associated with the entities from $Z$, $cn(Z)$, is then 
$cn(Z) = |\bigcup_{x \in Z} C^x_?|$ where $?$ is one of $T, S$ (depending on 
whether we are interested in the top or specific clusters, respectively). The 
size of a cluster $C_i \in C^x_?$, $cs(C_i)$, is an absolute frequency of the 
mentions of $C_i$ among the clusters associated with the entities in $Z$. More 
formally, $cs(C_i) = |\{x | x \in Z \wedge C_i \in C^x_?\}|$. 

The taxonomies can be used for defining {\bf taxonomy-based similarity} that 
reflects the closeness of entities depending on which clusters they belong to. 
The similarity is
$
sim_{tax}(x,y) = max(\{\frac{2 \cdot dpt(lcs(u,v))}{dpt(u)+dpt(v)} |$ 
$u \in C^x_S, v \in C^y_S \}),
$
where the specific tree codes in $C^x_S, C^y_S$ are interpreted as nodes
in the taxonomy induced by the dataset's hierarchical clustering. The $lcs$ 
function computes the least common subsumer of two nodes in the taxonomy and 
$dpt$ is the depth of a node in the taxonomy (defined as zero if no node is 
supplied as an argument, \ie if $lcs$ has no result). The formula we use is 
essentially based on a popular Wu-Palmer similarity 
measure~\cite{10.1371/journal.pcbi.1000443}. We only maximise it across all 
possible cluster annotations 
to find the best match (as the data are supposed to be unambiguous, such a 
strategy is safe). 
To illustrate the taxonomy-based similarity, let us assume the hierarchical
clusters from the previous example: \{{\tt C3.C1}\} for nodes {\tt A, D}; 
\{{\tt C3.C2}\} for node {\tt C}; \{{\tt C3.C1, C3.C2}\} for node {\tt B}. 
The taxonomy-based similarities between the nodes are then as follows:
$sim_{tax}(\mathtt{A,C}) = sim_{tax}(\mathtt{D,C}) = 0.5$ (the taxonomy root
is their least common subsumer), $sim_{tax}(\mathtt{B,C}) = 
sim_{tax}(\mathtt{B,A}) = sim_{tax}(\mathtt{B,D}) = sim_{tax}(\mathtt{A,D}) = 
1$ (the nodes are siblings). 

\subsection{Heuristic Quasi-Random Walks}\label{sec:methods.walks}

The \acronym~measures of a dataset are computed using its 
$G_w$ representation on which we execute multiple {\bf 
heuristic quasi-random walks} defined as follows. 
Let $l$ be a 
natural number and $h: V \rightarrow V$ a heuristic function that selects a
node to follow for any given node in $G_w$. Then a heuristic quasi-random 
walk on $G_w$ of length $l$ according to heuristic $h$ is an ordered tuple 
$W = (v,h(v),h(h(v)),\dots,$ $h^{l-1}(v),h^l(v))$ where $v$ is a random 
initial node in $G_w$. 
The walks simulate exploration of RDF datasets, either by a human user 
browsing the corresponding graph, or by an automated traversal and/or query 
agent. We use the indirect representation to cater for a broader range of 
possible traversal strategies (agents can easily explore the 
subject-predicate-object links in both directions, for instance by means of 
describe queries). 
By running a high number of walks, one can examine characteristic 
patterns of the dataset much earlier then by an exhaustive exploration of all 
possible connections (which is generally in the $O(n!)$ range w.r.t. the 
number of entities). Formal bounds of representativeness implied by a specific 
number of random walks are currently an open problem. However, our experiments 
suggest that a number ensuring 
representative enough sampling can be easily determined empirically. 

To simulate different types of exploration, we can define various 
{\bf heuristics}.  
For a given input node $v$, all 
heuristics compute a ranked list of the neighbours of $v$. The list is then 
iteratively processed (starting with the highest-ranking neighbour), 
attempting to select the next node with a probability that is inversely 
proportional to its rank. If no node has been selected after processing the 
whole list, a random neighbour is picked. 
The distinguishing factor of the heuristics are the criteria for ranking
the neighbour list. We employed the following selection preferences in our 
experiments:
\begin{inparaenum}[(1)]
  \item 
  nodes that have not been visited before (H1);
  \item 
  unvisited nodes connected by edges with 
  higher weight (H2);
  \item 
  unvisited nodes that are more similar to the
  current one, using the $sim_{tax}$ similarity introduced before (H3); 
  \item 
  unvisited nodes that are less similar (H4).
\end{inparaenum}
H1 simulates more or less random exploration that, however,
prefers 
unvisited nodes. H2 
follows
more significant relations. Finally, H3 and H4 are dual heuristics, with 
H3 simulating exploration 
of topics related to the current node
and H4 attempting to cover 
as many topics as possible.

Each walk $W$ can be associated with an {\bf envelope} $e(W,r)$ with a radius 
$r$, which is a sub-graph of $G_w$ limited to a set of nodes $V_W^r$. 
$V_W^r$ 
represents a neighbourhood of the walk and is defined as 
$\bigcup_{u \in W} \{v | v \in V \wedge |p_{G_w}(u,v)| \leq r\}$ where 
$p_{G_w}(u,v)$ is a shortest path between nodes $u,v$ in $G_w$. The envelope 
is used for computing the complexity and entropy measures later on, as it 
corresponds to the contextual information available to agents along a walk.

\subsection{\acronym~Measures}\label{sec:methods.measures}

Having introduced all the preliminaries, we can finally 
define the measures used in our sample implementation of
\acronym. The first type of measures is based on {\bf complexity} of the graph 
representations. We distinguish between local and global complexities. The 
global ones are associated with the graphs as a whole, and we compute 
specifically graph diameters, average shortest paths and node distributions 
along walks. The local measures associated with the walk envelopes 
are:
\begin{inparaenum}[(A)]
  \item envelope size in nodes;
  \item envelope size in biconnected components;
  \item average component size in nodes;
  \item average clustering coefficient of the walk nodes w.r.t. the envelope 
  graph. 
\end{inparaenum}

The {\bf coherences} of walks are based on similarities. 
Let us assume a sequence of $v_1,v_2,\dots,v_n$ walk nodes. Then the particular
coherences are:
\begin{inparaenum}[(A)]\setliststart{5}
  \item taxonomy-based start/end coherence $sim_{tax}(v_1,v_n)$;
  \item taxonomy-based product coherence $\Pi_{i \in \{1,\dots,n-1\}}$ 
  $sim_{tax}(v_i,v_{i+1})$;
  \item average taxonomy-based coherence  
  $\frac{1}{n-1}$ $\sum_{i \in \{1,\dots,n-1\}}$ $sim_{tax}(v_i,v_{i+1})$;
  \item distributional start/end coherence $sim_{cos}(v_1,$ $v_n)$;
  \item distributional product coherence $\Pi_{i \in \{1,\dots,n-1\}} 
  sim_{cos}(v_i,v_{i+1})$;
  \item average distributional coherence 
  $\frac{1}{n-1}\sum_{i \in \{1,\dots,n-1\}} sim_{cos}(v_i,v_{i+1})$.
\end{inparaenum}
This family of measures helps us to assess how topically convergent (or 
divergent) are the walks. 

To compute walk {\bf entropies}, we use the $T_w, T_s$ taxonomies. By 
definition, the higher the entropy of a variable, the more information the 
variable contains. In our context, a high entropy value associated with a walk 
means that there is a lot of information available for agents to possibly 
utilise when processing the graph. The entropy measures we use relate to the 
following sets of nodes and types of clusters representing the context of the 
walks:
\begin{inparaenum}[(A)]\setliststart{11}
  \item walk nodes only, top clusters;
  \item walk nodes only, specific clusters;
  \item walk and envelope nodes, top clusters;
  \item walk and envelope nodes, specific clusters. 
\end{inparaenum}
The entropies of the sets (K-N) are defined using the notion of cluster size 
($cs(\dots)$) introduced before. Given a set $Z$ of nodes of interest, the 
entropy $H(Z)$ 
is computed as
$
H(Z) = - \sum_{C_i \in C_?(Z)} \frac{cs(C_i)}{\sum_{C_j \in C_?(Z)}cs(C_j)} 
         \cdot \log_2 \frac{cs(C_i)}{\sum_{C_j \in C_?(Z)}cs(C_j)},
$
where $?$ is one of $T, S$, for top or specific clusters, respectively. 

\subsection{Interpreting the Measures}\label{sec:methods.interpretation}

Generally speaking, high complexity means a lot of potentially useful 
structural information, but also more expensive search (\eg by means of 
queries) due to high branching
factors among the nodes, and the other way around. 
High coherence means that in
general, any exploratory walk through the dataset tends to be focused in 
terms of topics covered, while low coherence indicates rather serendipitous
nature of a dataset where exploration tends to lead through many different 
topics. Finally, high entropy means more information and also less predictable 
topic distributions along the nodes in the walks and envelopes, with balanced 
cluster cardinalities. Low entropy means high predictability of the node 
topics (in other words, strongly skewed cluster cardinalities). 

Possible combinations of measures can be enumerated 
as follows. Let us refer to comparatively higher and lower measures by the
\ua~and \da~symbols. Then the combinations of relative complexity, coherence
and entropy measures, respectively, are:
\begin{inparaenum}[1,]
  \item \ua\ua\ua: Complex patterns and informative topic annotations about 
  focused subject domains.
  \item \ua\ua\da: Focused around unevenly distributed sets of topics with 
  complex structural information context.
  \item \ua\da\ua: Serendipitous, a lot of equally significant complex 
  contextual information.
  \item \da\ua\ua: Focused, with balanced and simple contextual information. 
  \item \ua\da\da: Serendipitous with complex contextual topics of uneven 
  cardinality.
  \item \da\ua\da: Focused with simple uneven contexts.
  \item \da\da\ua: Serendipitous with simple balanced contexts.
  \item \da\da\da: Serendipitous with simple uneven contexts.
\end{inparaenum}

Some of the specific measure combinations may be particularly (un)suitable for
certain use cases. To give few non-exhaustive examples, the 
combination 1, \ua\ua\ua~is suitable for knowledge discovery about focused
subject domains, but also challenging for querying. Combination 3, \ua\da\ua~is
good for serendipitous browsing. Combination 4, \da\ua\ua~may be useful 
for semantic annotations of a set of core domain entities as it provides for 
simple lookups of focused and balanced contextual information. Similarly, 
combination 7, \da\da\ua~may be more applicable for annotations of varied
domain entities. 


\section{Experiments}\label{sec:experiments}

In this section, we first present settings of experiments with 
\acronym~applied to sample RDF datasets. Then we report on results of the 
experiments and discuss their interpretation. Note that the 
implementation of the \acronym~methodology used in the experiments, including 
the corresponding data and scripts, is available at \url{http://goo.gl/Wxnb3B}.

\subsection{Datasets and Settings}\label{sec:experiments.settings}

The datasets we used were:
\begin{inparaenum}[1.]
  \item {\it DrugBank} -- information on marketed drugs, including indications,
  chemical and molecular features, manufacturers, protein bindings, etc.;
  \item {\it SIDER} -- information on drug side effects;
  \item {\it Diseasome} -- a network of disorders and associated genes;
  \item {\it all} -- an aggregate of the {\it DrugBank}, {\it SIDER} and 
  {\it Diseasome} datasets using the DrugBank URIs as a core vocabulary to 
  which the other datasets are mapped.
\end{inparaenum}
The dataset selection was motivated by our recent work in adverse drug effect
discovery, for which we have been compiling a knowledge base from relevant
biomedical Linked Open Data~\cite{ahmed2014amia}. One of the main purposes of
the knowledge base is to extract features applicable to training adverse 
effect discovery models. In this context, we were interested in characteristics
of the knowledge bases corresponding to the isolated and merged datasets, yet 
we lacked the means for measuring this. Therefore we decided to use the 
knowledge bases being created in~\cite{ahmed2014amia} as a test case for 
\acronym.

For each dataset, we generated:
\begin{inparaenum}[(1)]
  \item The direct graph and distributional representations $G_d, M$, with $M$
  reduced to $250$ most significant dimensions according to their $\chi^2$ 
  scores.
  \item The weighted indirect and similarity representations $G_w, G_s$, 
  taking into account only similarity values above $0.5$.
  \item Taxonomies $T_w, T_s$ based on the $G_w, G_s$ graph clustering, 
  respectively.
\end{inparaenum}

The quasi-random heuristic walks were ran using all combinations of the 
following parameters for each dataset:
\begin{inparaenum}[(1)]
  \item Walk lengths $l \in \{2, 10, 20\}$.
  \item Envelope diameters $r \in \{0,1\}$.
  \item Heuristics $h \in \{H1, H2, H3, H4\}$ (\ie random, weight, similarity 
  and dissimilarity preference). 
\end{inparaenum}
The number of samples (\ie walk executions per a parameter combination) was 
$\frac{|V|}{k(l+1)}$, where $|V|, l$ are the number of graph nodes and the walk 
length in the given experimental batch, respectively, and $k$ is a constant 
equal to the average shortest path length in the graphs, truncated to integer 
value. In our experiments, the observed relative trends were stable after 
reaching this number of repetitions and therefore we took it as a sufficient 
`sampling rate.' 

\subsection{Results}\label{sec:experiments.results}

\begin{figure}[ht]
\center
\scalebox{0.19}{\includegraphics{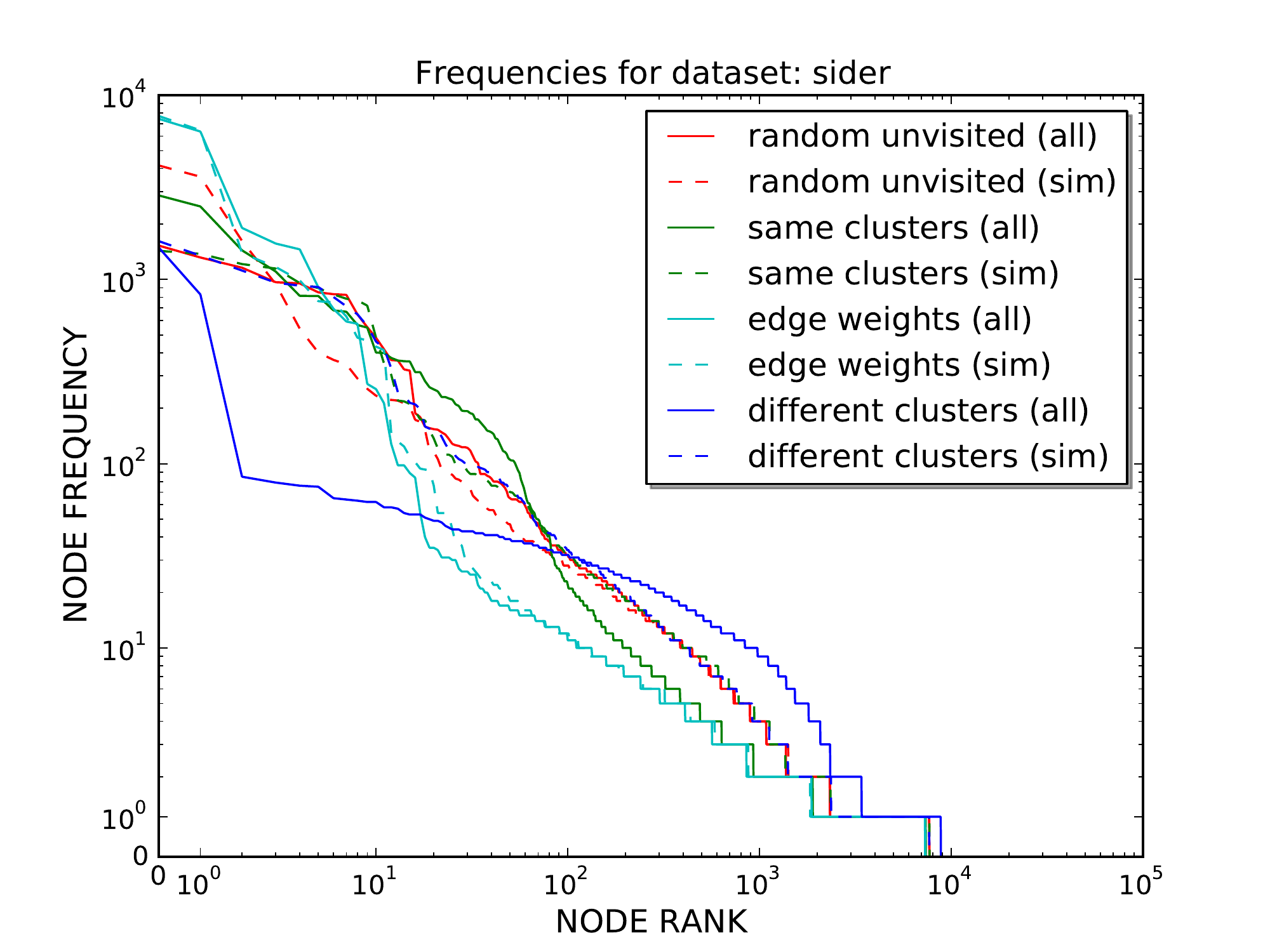}}
\scalebox{0.19}{\includegraphics{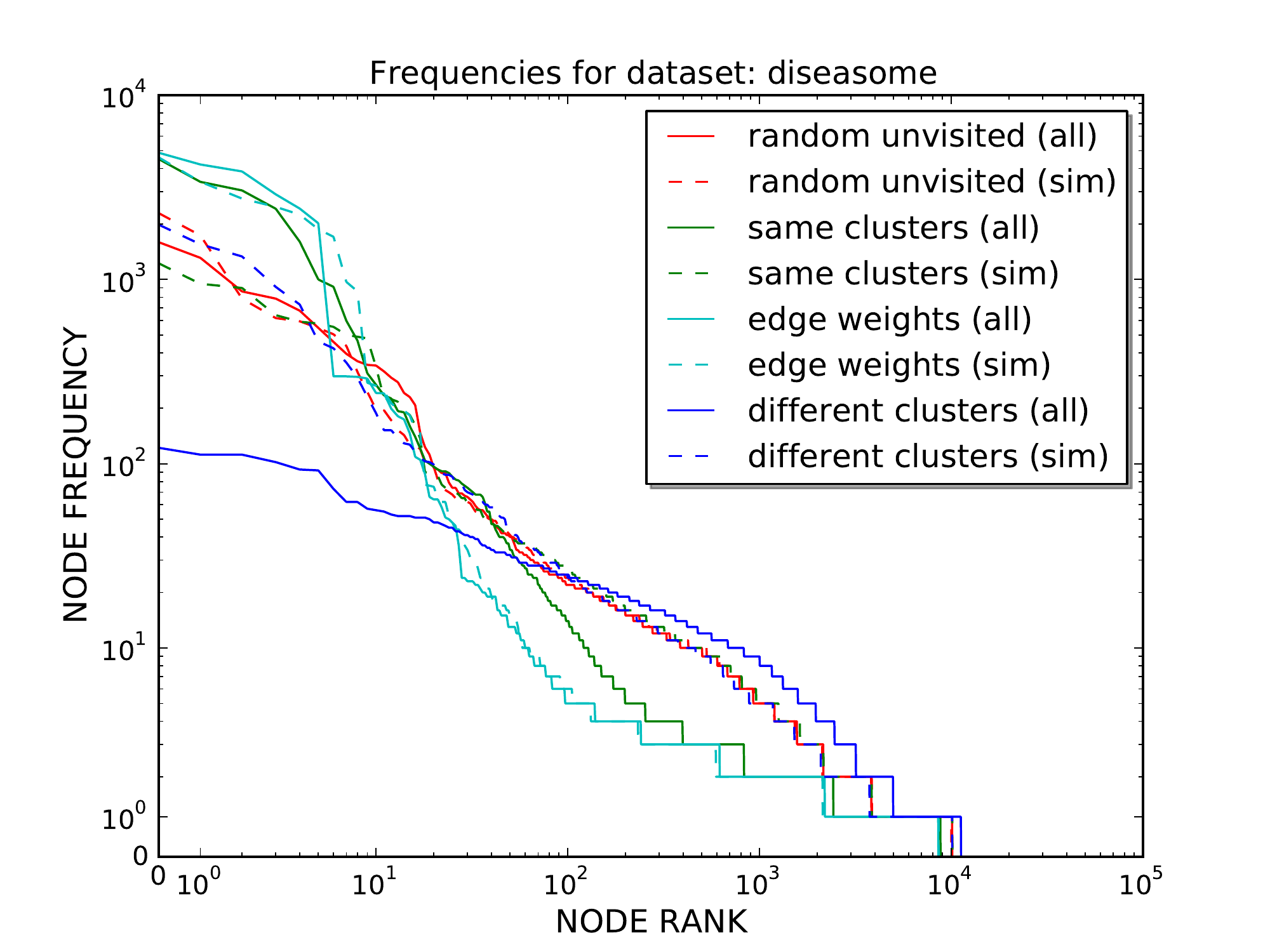}}\\
\scalebox{0.19}{\includegraphics{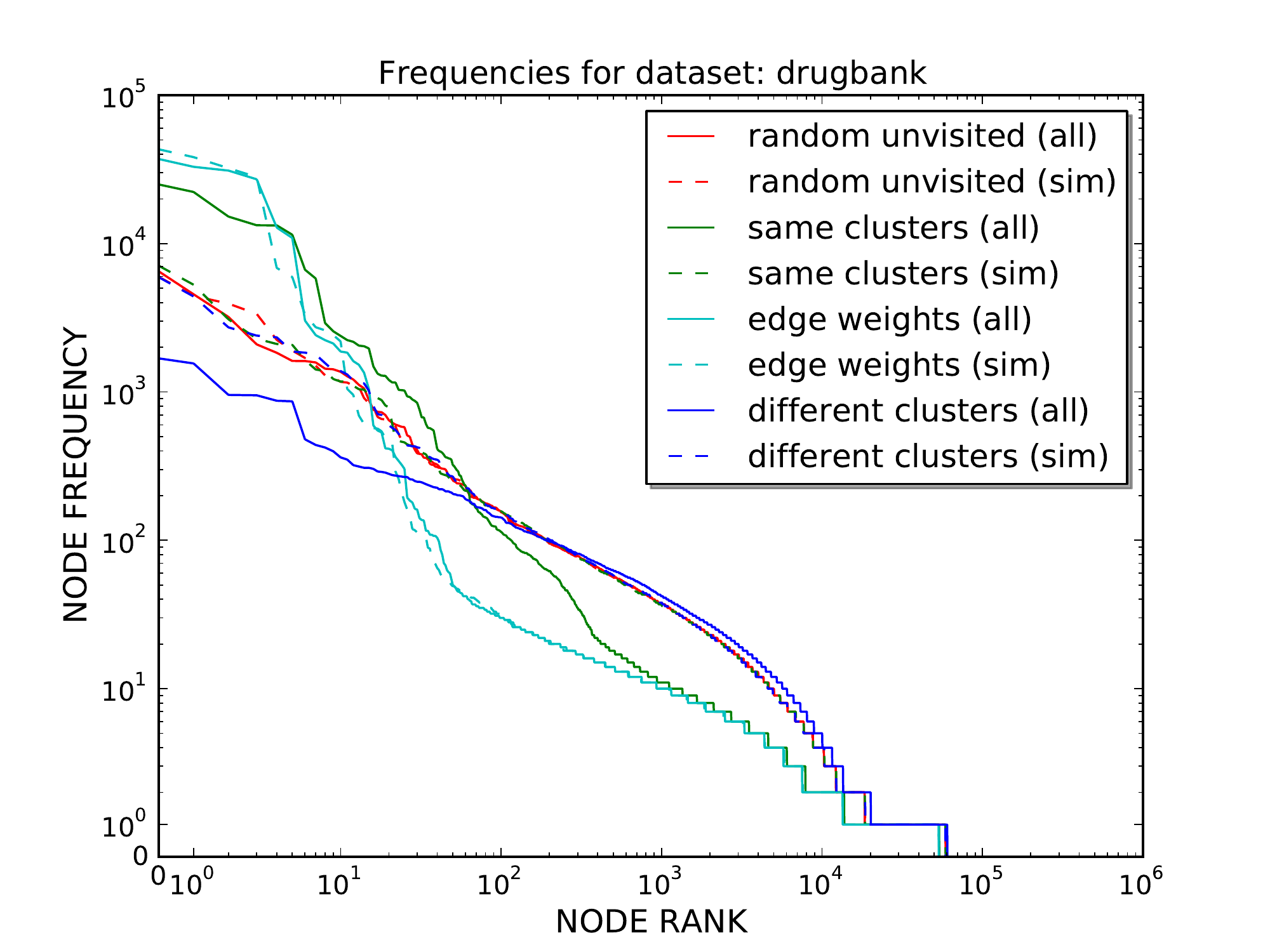}}
\scalebox{0.19}{\includegraphics{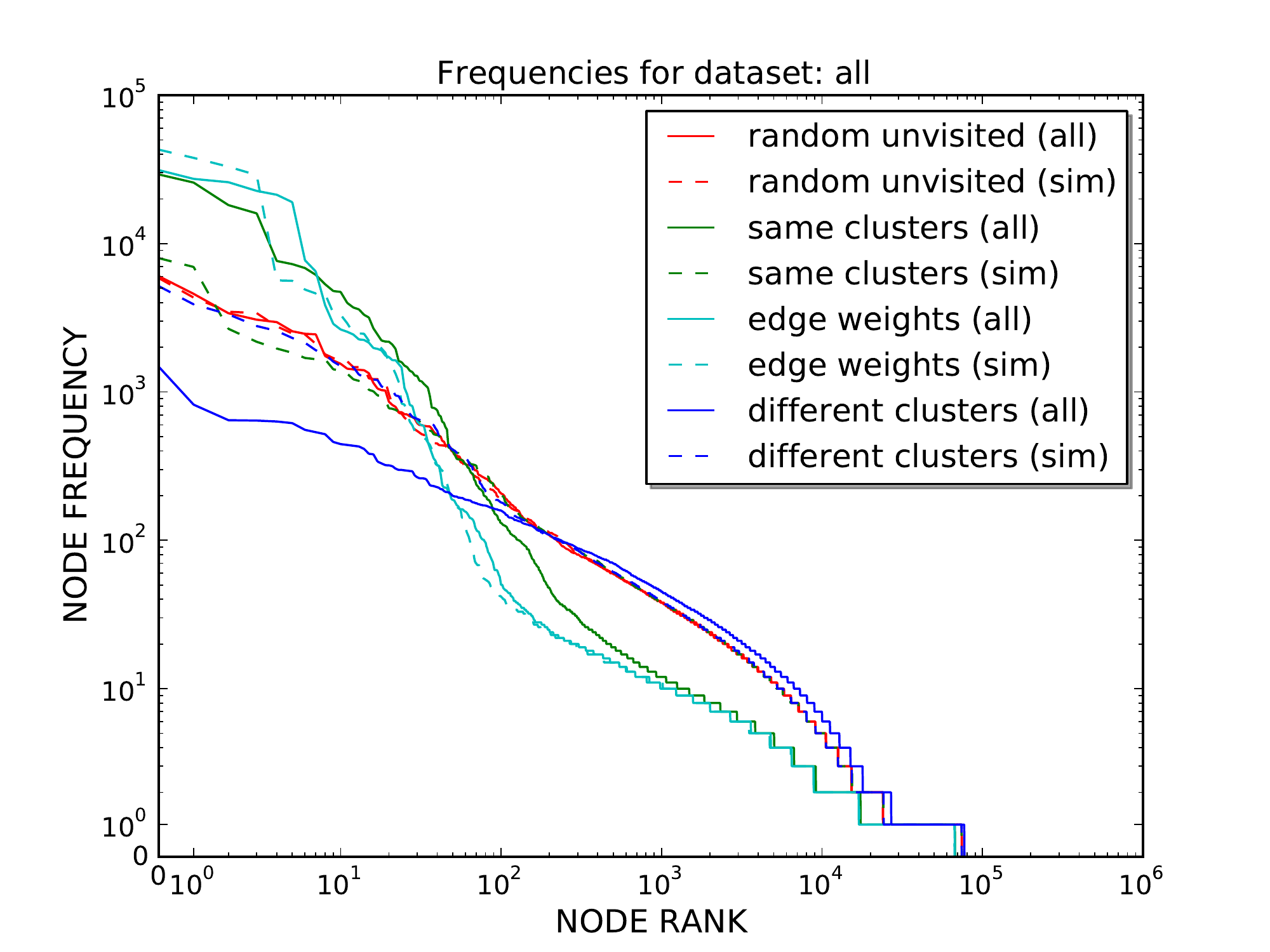}}
\caption{Node distributions along the walks}\label{fig:node_dist}
\end{figure}

Figure~\ref{fig:node_dist} gives an overview
of how the specific heuristics perform per each dataset regarding the {\bf node 
visit frequency}. 
\vspace{-0.3cm}
\vspace{-0.3cm}
The x-axis reflects the ranking of nodes according to the number of visits to
them. The y-axis represents the visit frequencies. Both axes are log-scale,
since all the distributions have very steep long tails.
The prevalent trends in the plots are:
\begin{inparaenum}[1.]
  \item The heuristics H2 and H3 (edge weight and similarity preference), 
  especially when using the $T_w$ taxonomies, 
  tend to have generally more long-tail distributions than 
  the others (the pattern is most obvious in the {\it Diseasome} dataset).
  \item The H4 heuristic, using the $T_w$ taxonomy, has the most even
  distribution.
  \item The heuristics using the $T_s$ taxonomies tend to have very similar
  node visit frequency distributions, close to H1 that exhibits the most
  `average' behaviour (presumably due to its highest randomness).
  \item The heuristics seem to follow similar patterns in the {\it DrugBank} 
  and {\it all} datasets.
  \item In {\it SIDER}, the behaviour of the heuristics appears to be most 
  irregular (for instance, the random heuristic H1 behaves differently for
  $T_w$ and $T_s$ taxonomies although the taxonomy used should not have any 
  influence on that heuristic).
\end{inparaenum}

Table~\ref{tab:graph_stats} summarises {\bf global characteristics} 
of the datasets and the corresponding $G_w$ graph representations. 
\vspace{-0.5cm}
\begin{table}[ht]
\center
{\scriptsize
\begin{tabular}{|l||c|c|c|c|c|c|c|c|}
\hline
Data set ID & $|V|$ & $|E|$ & $\frac{|E|}{|V|}$ & $D$ & $d$ & $l_{G}$ & $|C|$ \\
\hline\hline
SIDER & $27,924$  & $96,427$  & $3.453$ & $0.000247$ & 
      $6.998$ & $4.385$ & $2$\\
\hline
Diseasome & $28,102$  & $64,172$  & $2.284$ & $0.000163$ & 
      $4.999$ & $3.914$ & $3$\\
\hline
DrugBank & $219,513$  & $361,389$ & $1.646$ & $0.000015$ & 
     $5.999$ & $4.352$ & $2$ \\
\hline
All & $265,548$ & $513,326$ & $1.933$ & $0.000015$ & 
       $7.998$ & $4.667$ & $3$ \\
\hline
\end{tabular}}
\caption{Global graph statistics}\label{tab:graph_stats}
\end{table}
\vspace{-1cm}
$|V|, |E|$ are numbers of nodes and edges in $G_w$, respectively, $D$ is the 
graph density (defined as $D = \frac{2 \cdot |E|}{|V|(|V|-1)}$), 
$d$ is the graph diameter, $l_{G}$ is the average
shortest path length and $|C|$ is the number of connected components. All
graphs have so called small world property~\cite{sna}, as their densities are
rather small and yet there is very little separation between any two nodes in
the graph in general. This typically happens in highly complex graphs with a 
lot of interesting patterns in them. 

\begin{figure}[ht]
\center
\scalebox{0.19}{\includegraphics{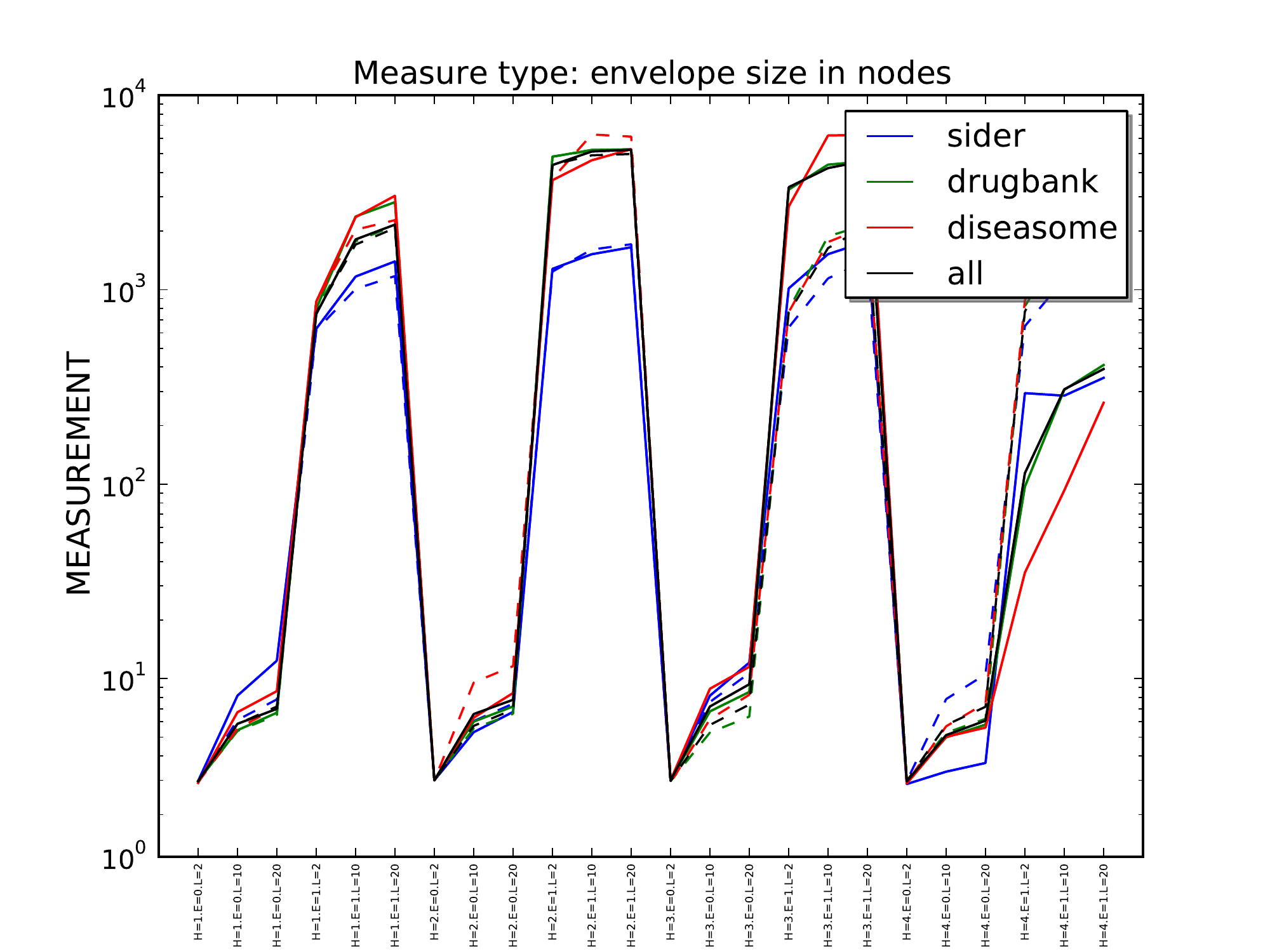}}
\scalebox{0.19}{\includegraphics{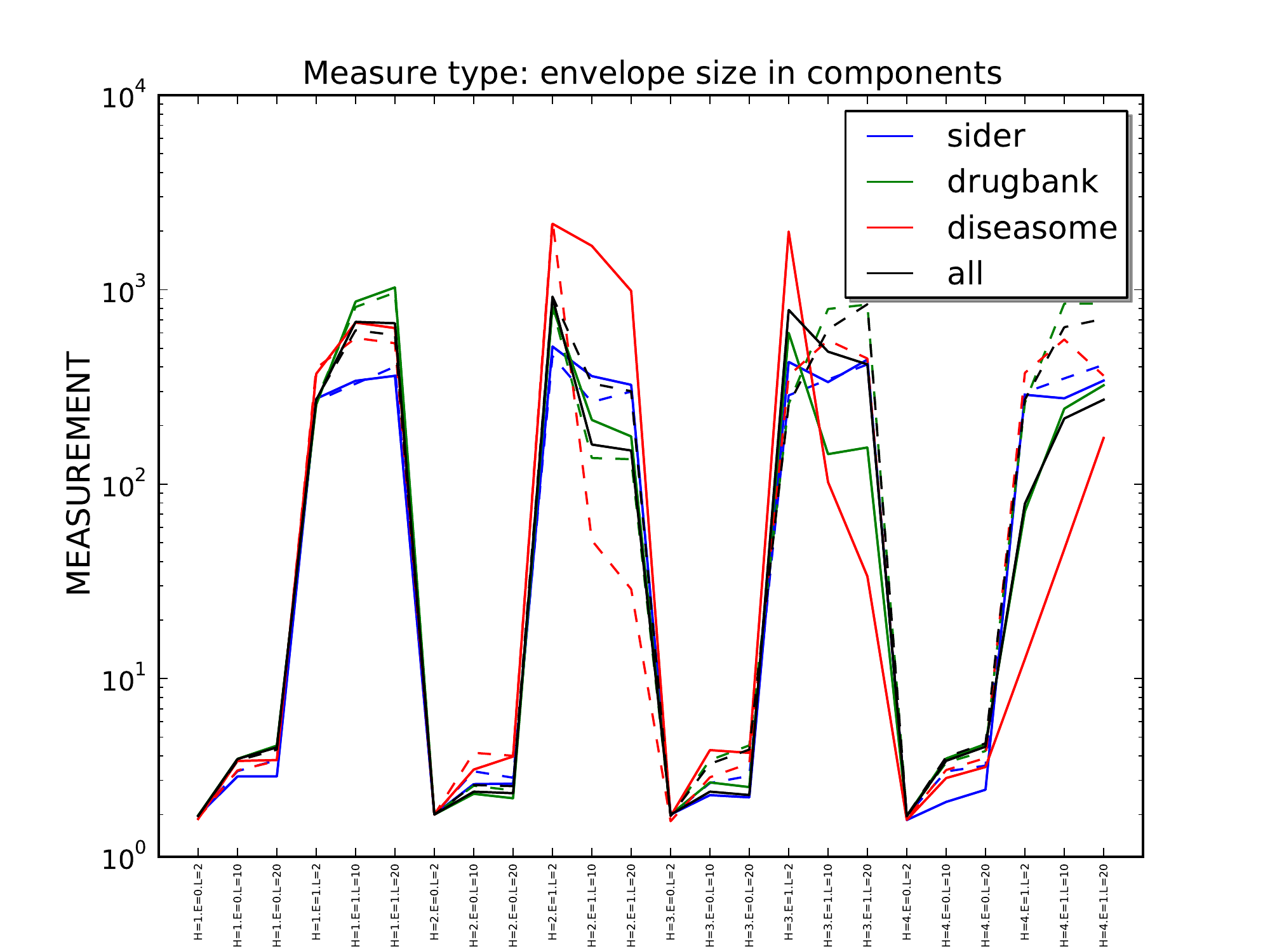}}\\
\scalebox{0.19}{\includegraphics{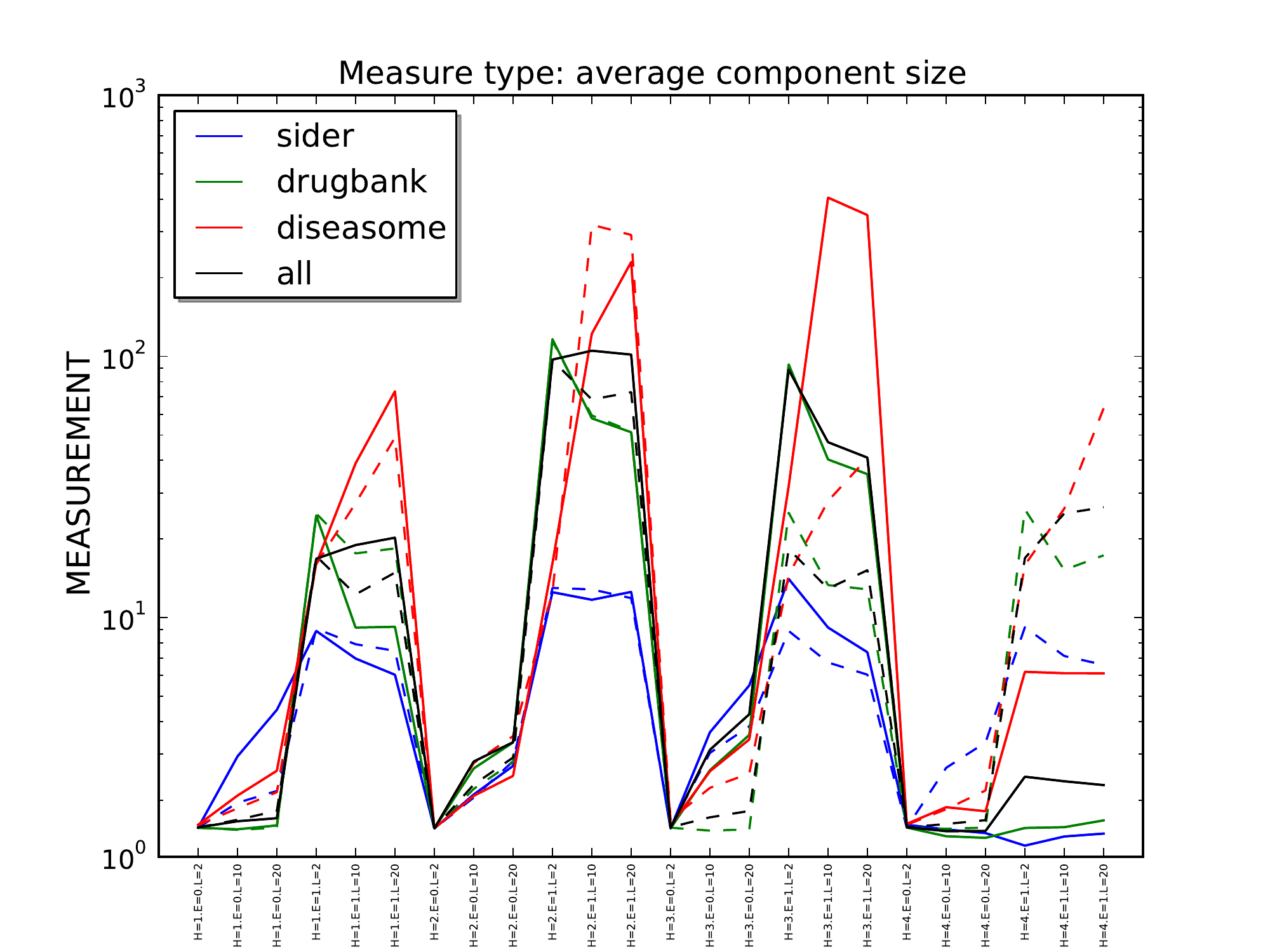}}
\scalebox{0.19}{\includegraphics{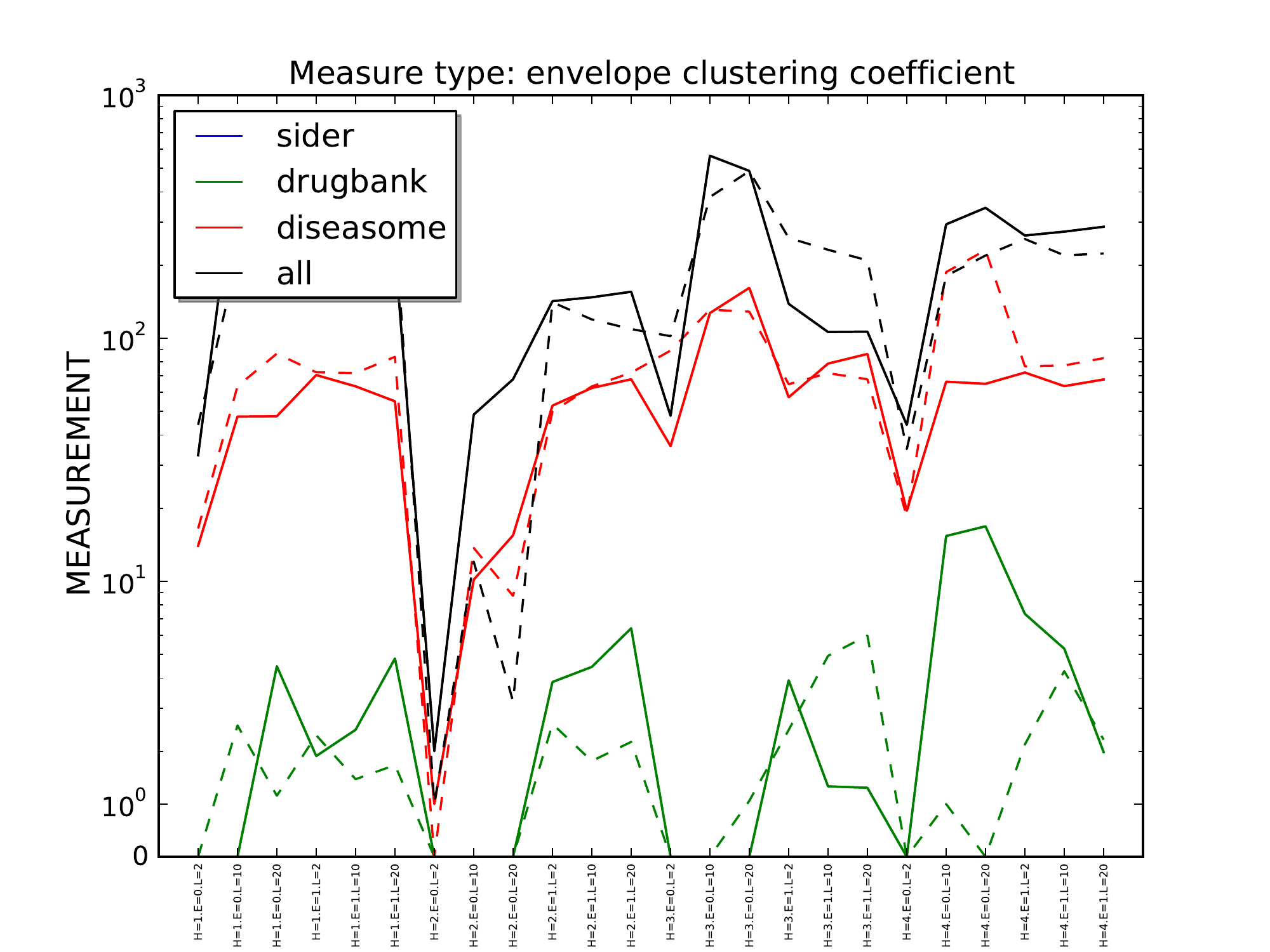}}
\caption{Complexity plots}\label{fig:complexity}
\end{figure}

Figure~\ref{fig:complexity} presents plots of the {\bf complexity} measures 
based on the walk sampling. 
\vspace{-0.3cm}
\vspace{-0.3cm}
The x-axis represents the combinations of experimental parameters, grouped
by the type of heuristic -- the 1.,2.,3. and 4. horizontal quarters of the 
plot correspond to H1, H2, H3 and H4, respectively. 
For each heuristic, there are six different
combinations of the path length and envelope diameter, progressively increasing
from left to right. The y-axis represents the actual value of the measure
plotted, 
rendered in an appropriate log-scale if there are too big relative differences 
between the plotted values. Each plot 
represents one type of measure and different colours correspond to specific
datasets (red for {\it Diseasome}, green for {\it DrugBank}, blue for 
{\it SIDER} and black for {\it all}). The full and dashed lines are for 
experiments using the $T_w$ and $T_s$ taxonomies, respectively. 
All 
the walk-sampling results reported below 
are plotted in this fashion.

The results of the complexity measures can be summarised as follows:
\begin{inparaenum}[1.]
  \item 
  The size and number of components increase with
  longer walks and larger envelopes.
  \item The {\it SIDER} dataset has generally lowest number of components of 
  smallest size, while {\it Diseasome} is dominating in these measures.
  \item The {\it all} dataset has relatively large components in average, but
  there is less of them than in case of {\it Diseasome}.
  \item The {\it all} dataset has the largest complexity in terms of clustering
  coefficients, with {\it Diseasome} being closely second and {\it DrugBank}
  comparatively much smaller. {\it SIDER} has zero complexity according to
  the clustering coefficient. 
\end{inparaenum}

\begin{figure}[ht]
\center
\scalebox{0.19}{\includegraphics{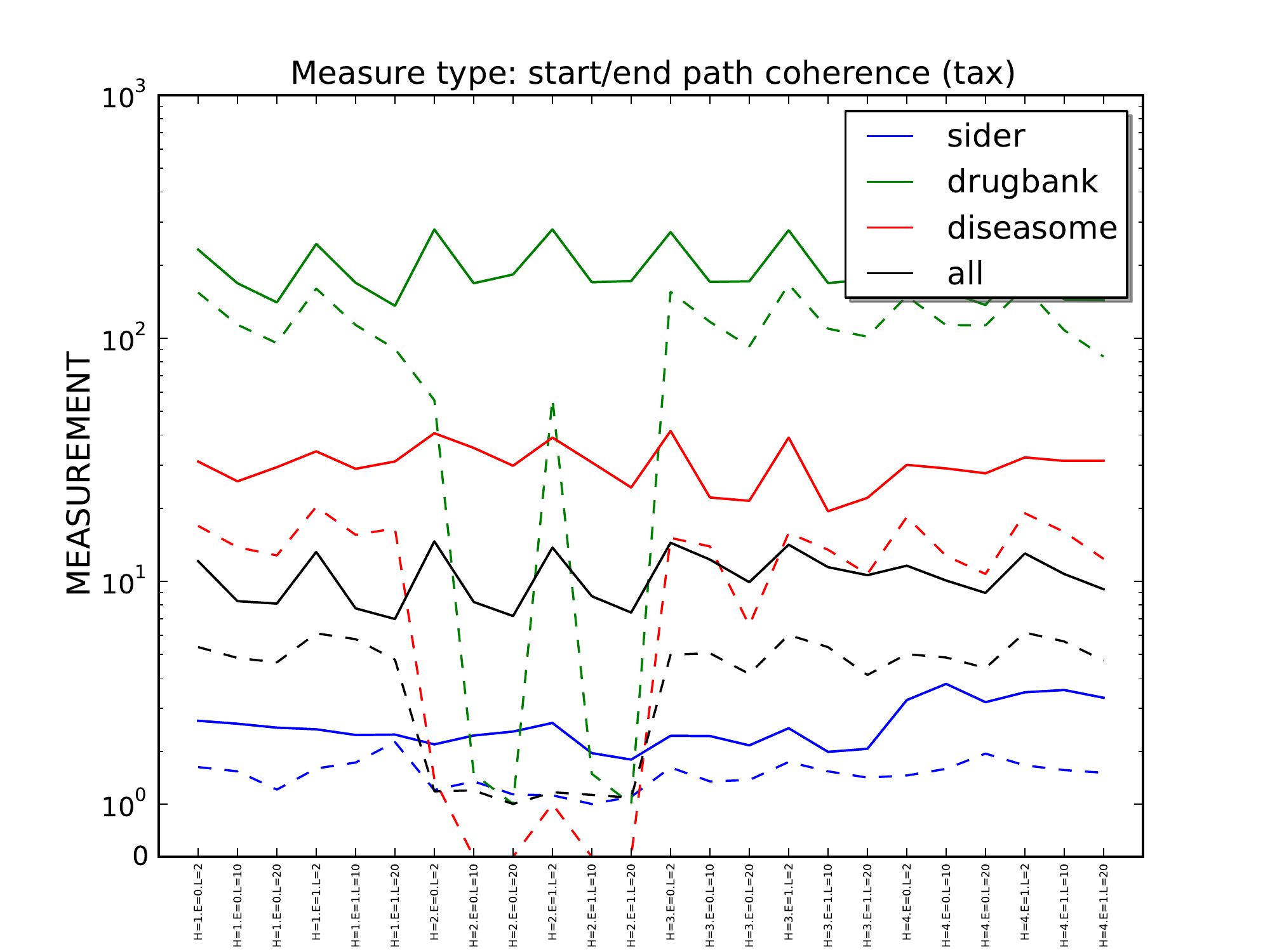}}
\scalebox{0.19}{\includegraphics{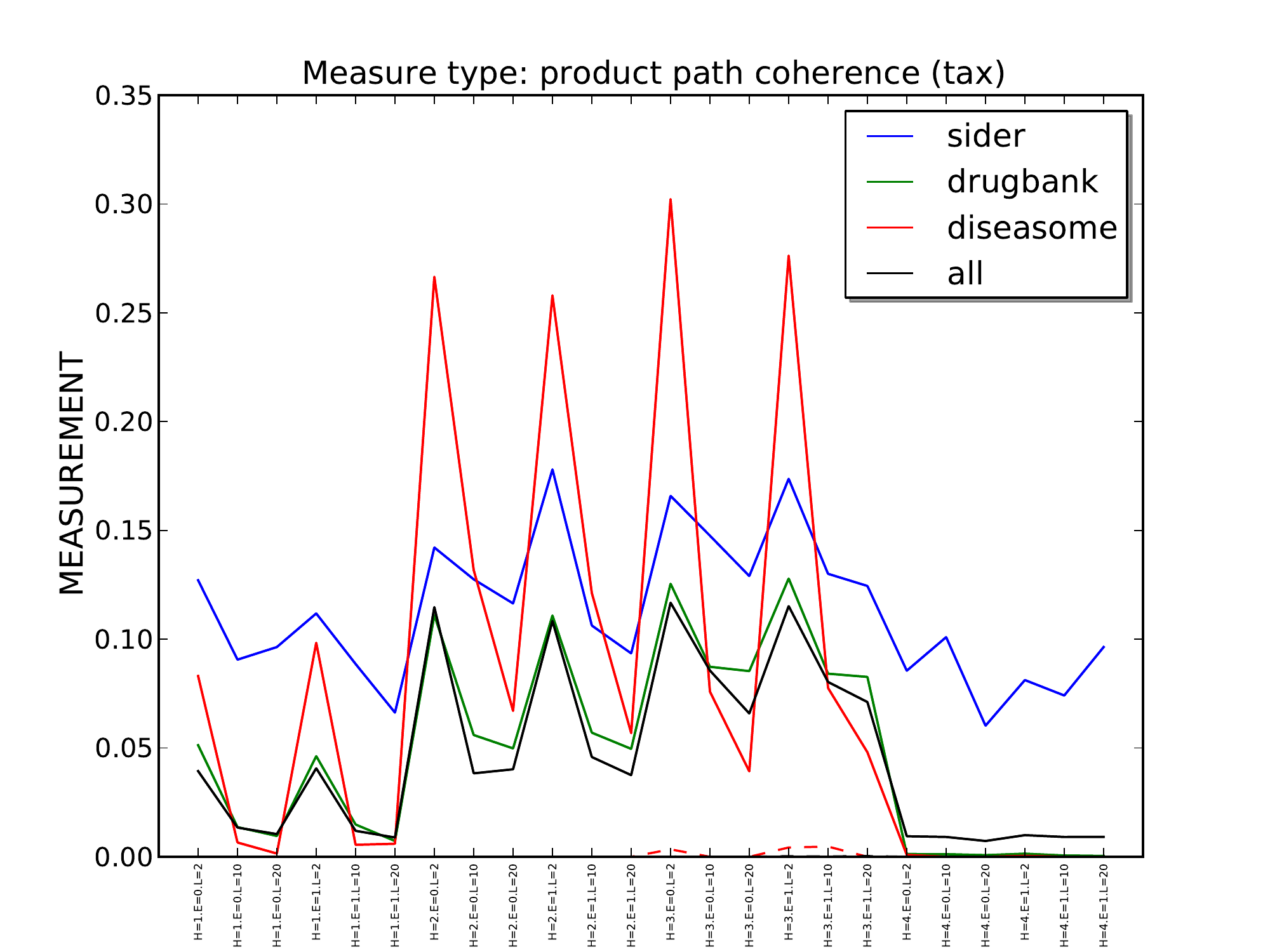}}
\scalebox{0.19}{\includegraphics{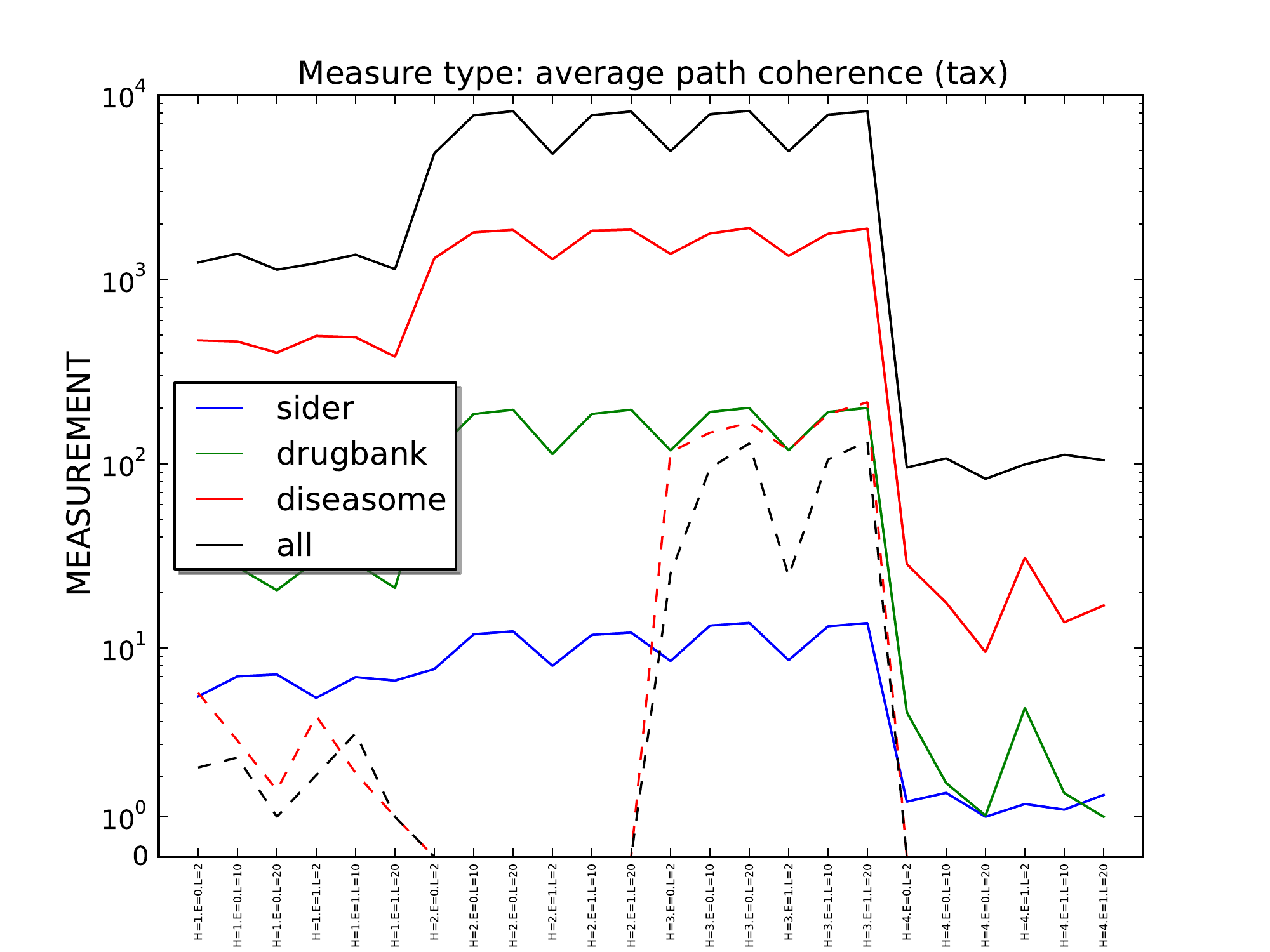}}\\
\scalebox{0.19}{\includegraphics{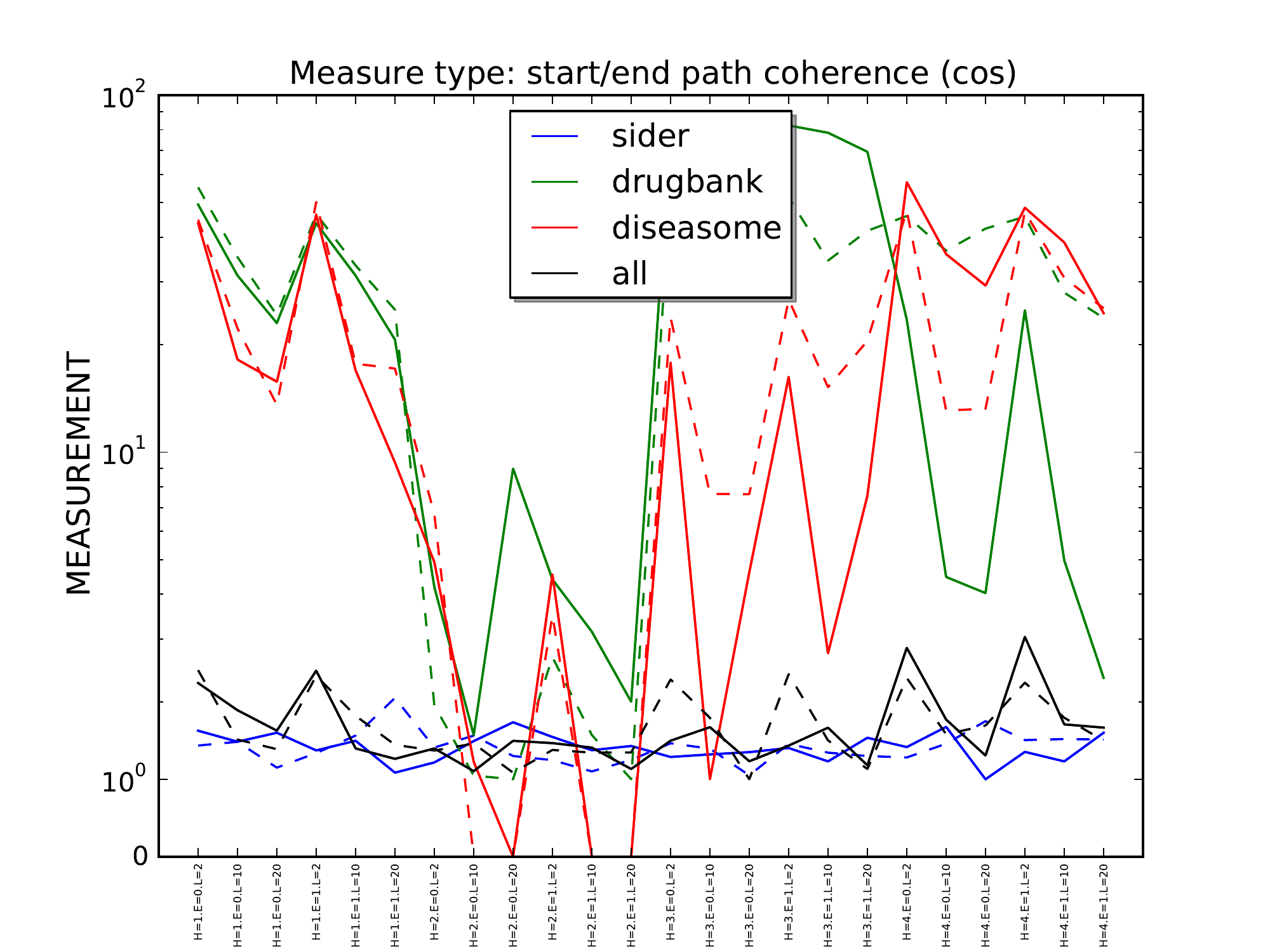}}
\scalebox{0.19}{\includegraphics{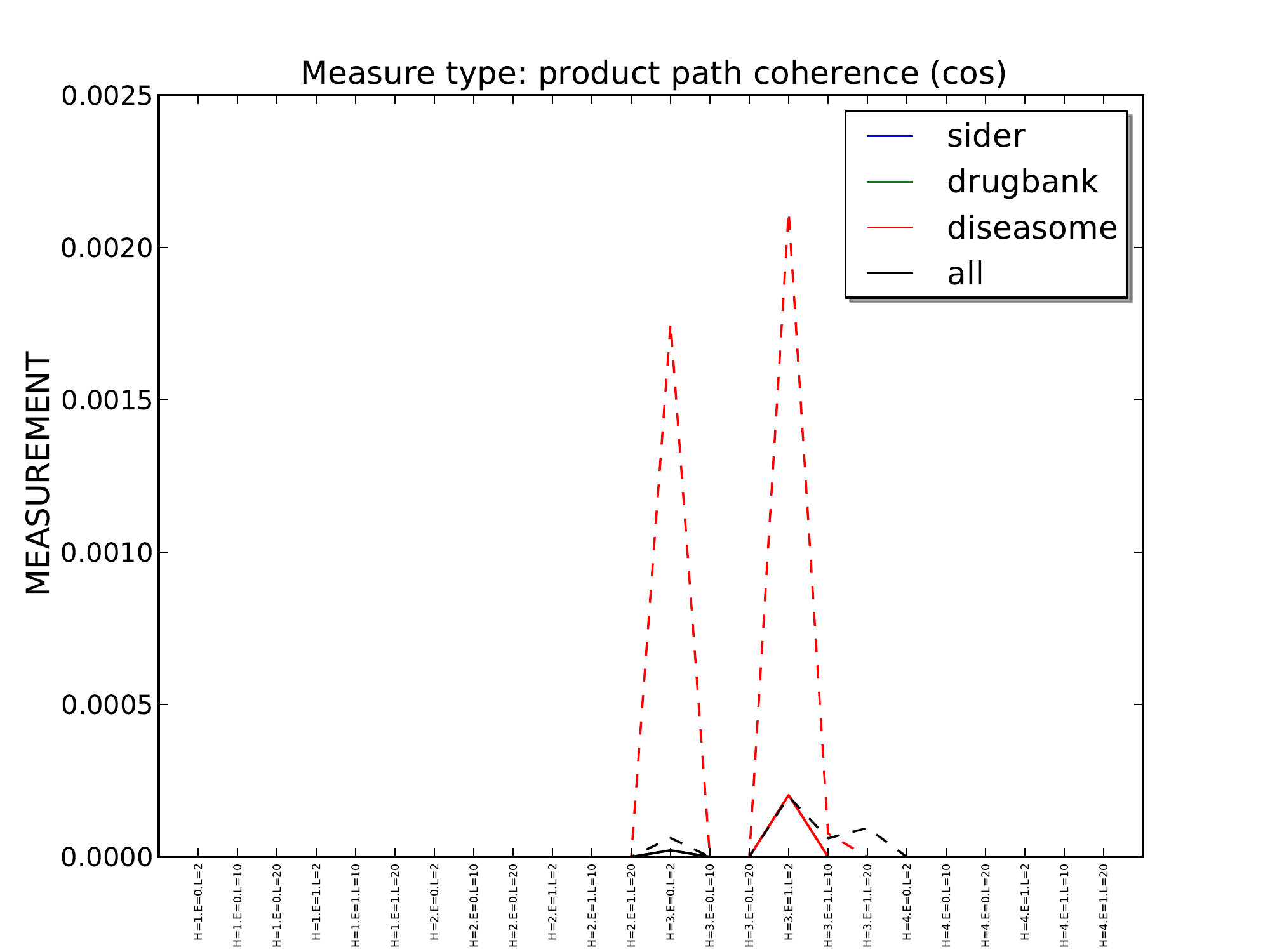}}
\scalebox{0.19}{\includegraphics{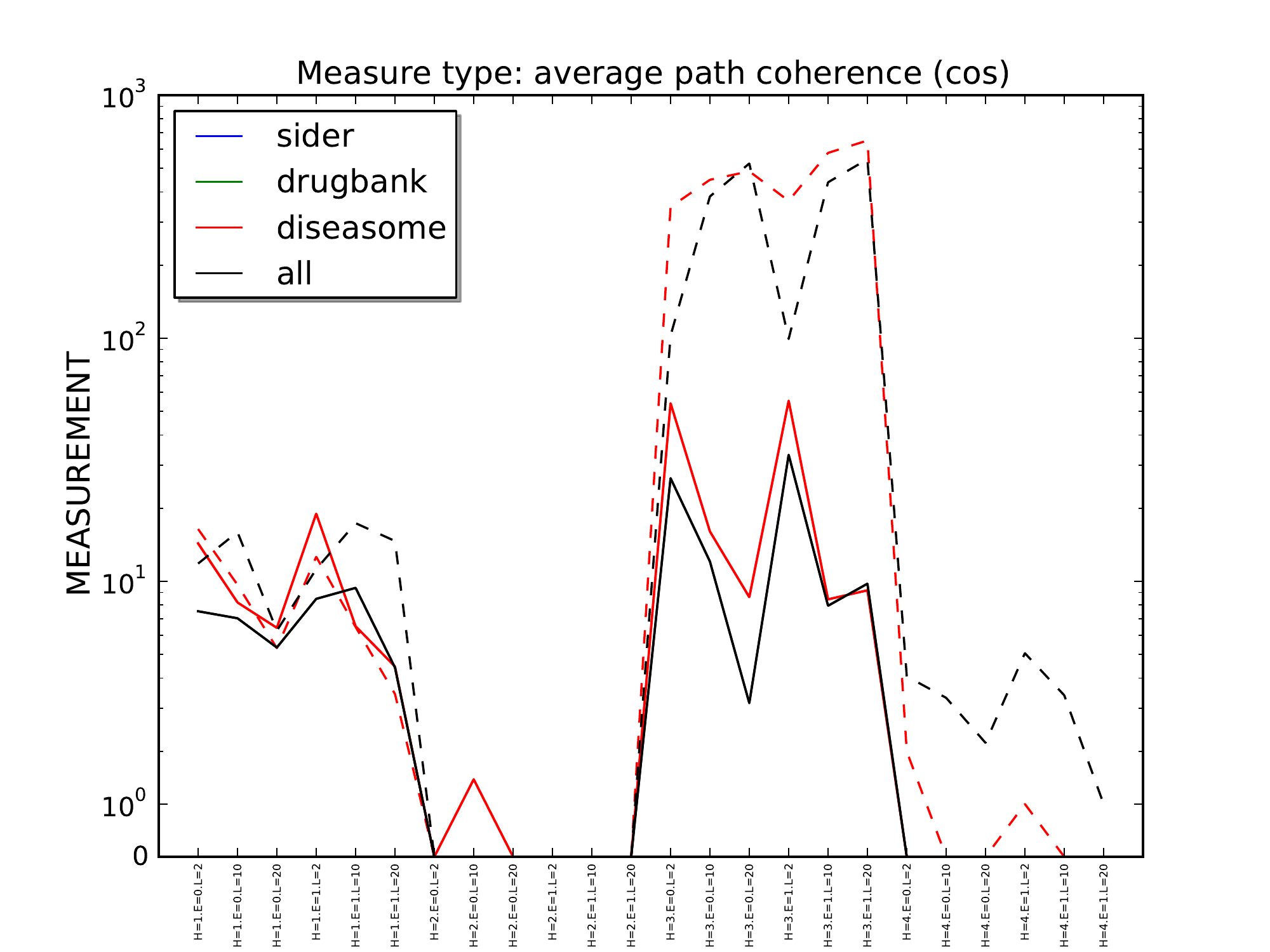}}
\caption{Coherence plots}\label{fig:coherence}
\end{figure}

The results of the {\bf coherence} analysis are in Figure~\ref{fig:coherence}.
\vspace{-0.3cm}
\vspace{-0.3cm}
The general observations are: 
\begin{inparaenum}
  \item The start/end coherences tend to be higher for shorter path lengths.
  \item The coherences in the samples using the $T_w$ taxonomies are generally
  higher than the ones using the $T_s$ taxonomy. 
  \item {\it SIDER} has the lowest coherence in most cases.
  \item The product and average coherences tend to be relatively 
  lower for the H4 (dissimilarity) heuristic. 
  \item The {\it Diseasome} dataset is generally the second best for most
  coherence types. {\it DrugBank} is generally third, except of the start/end
  coherence where it is mostly the best. For the average and product 
  coherences, 
  the {\it all} dataset usually performs best. The trend is clearer for the 
  coherences based on taxonomical 
  similarity. 
\end{inparaenum}

\begin{figure}[ht]
\center
\scalebox{0.19}{\includegraphics{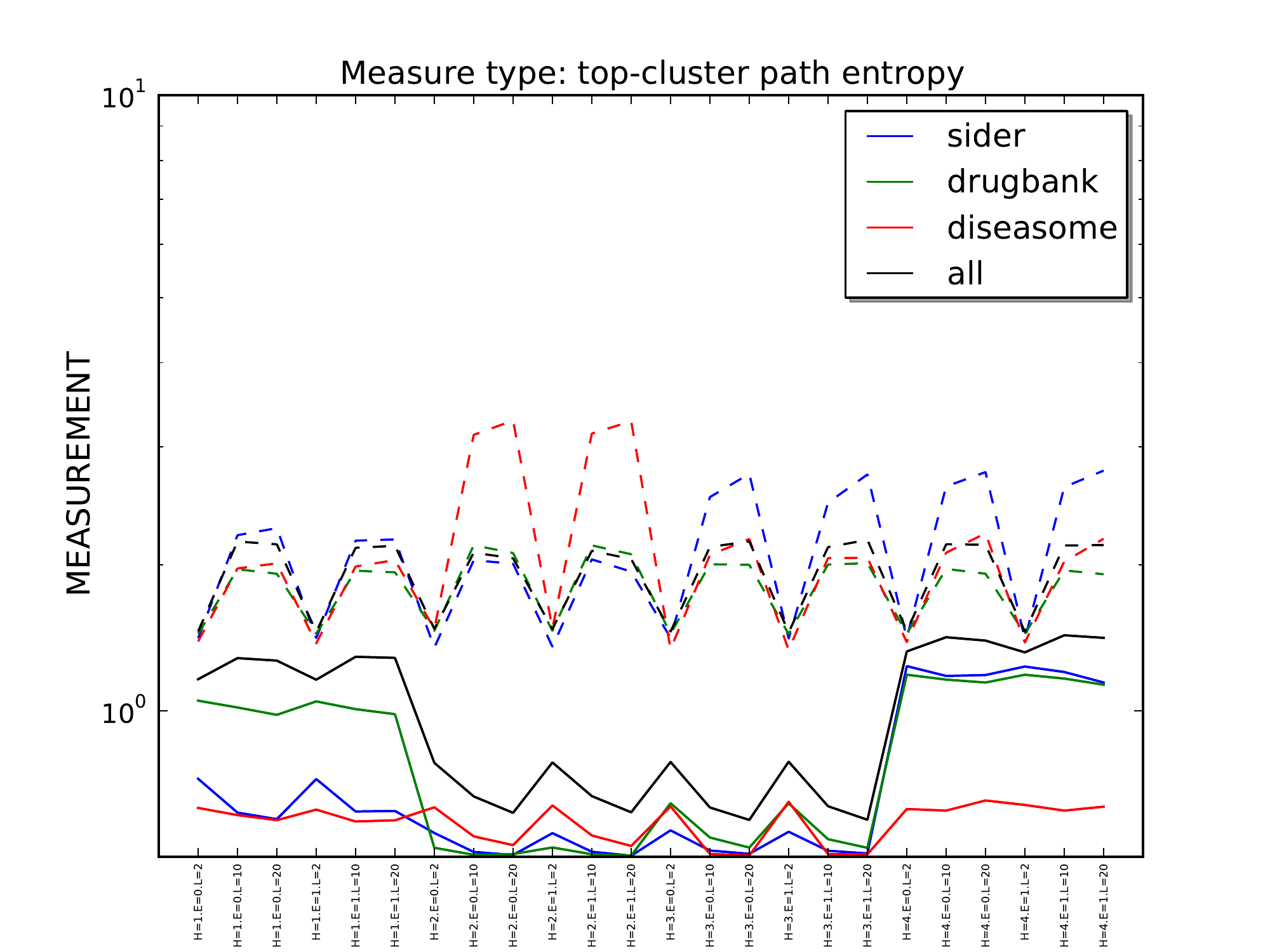}}
\scalebox{0.19}{\includegraphics{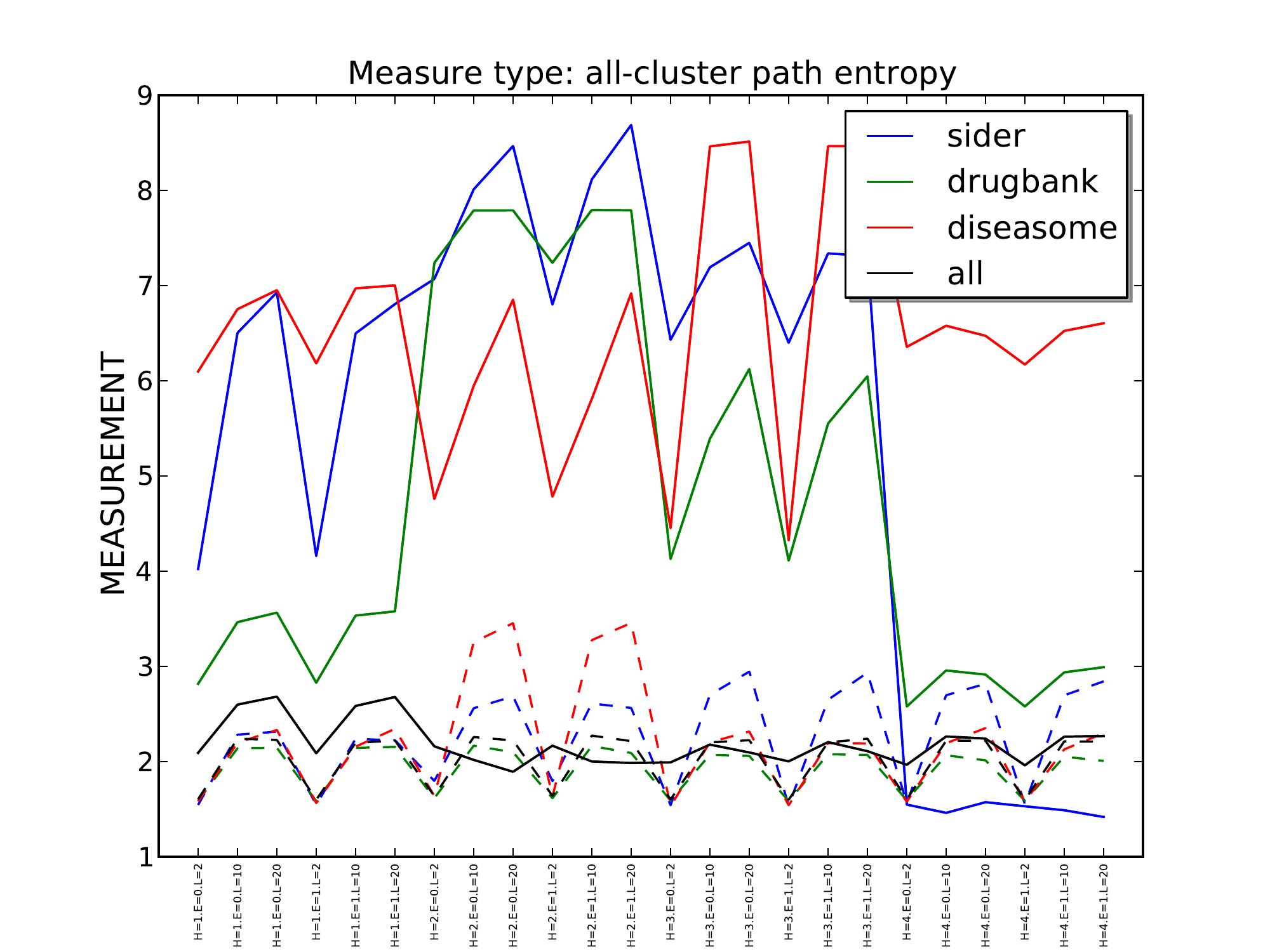}}\\
\scalebox{0.19}{\includegraphics{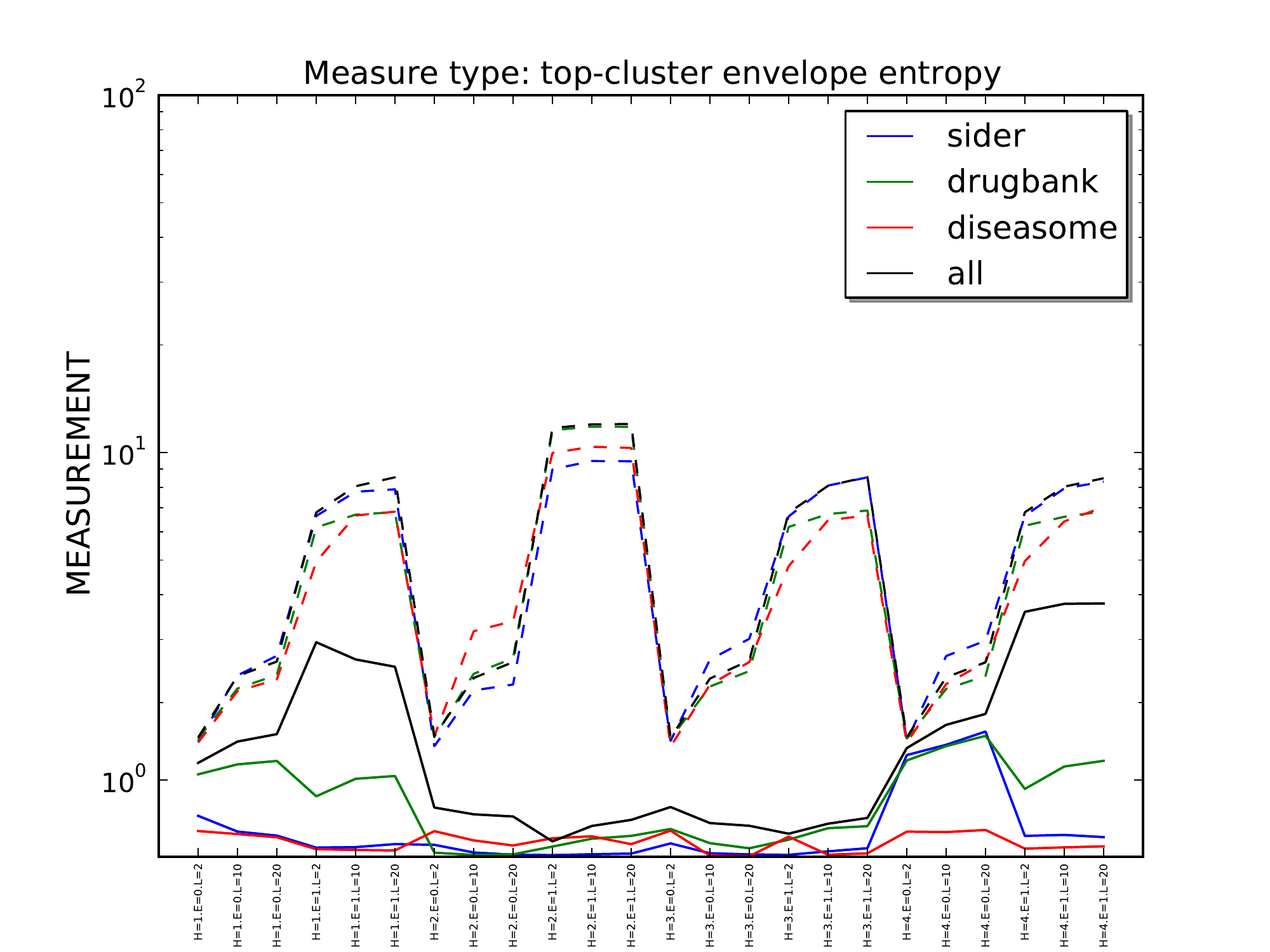}}
\scalebox{0.19}{\includegraphics{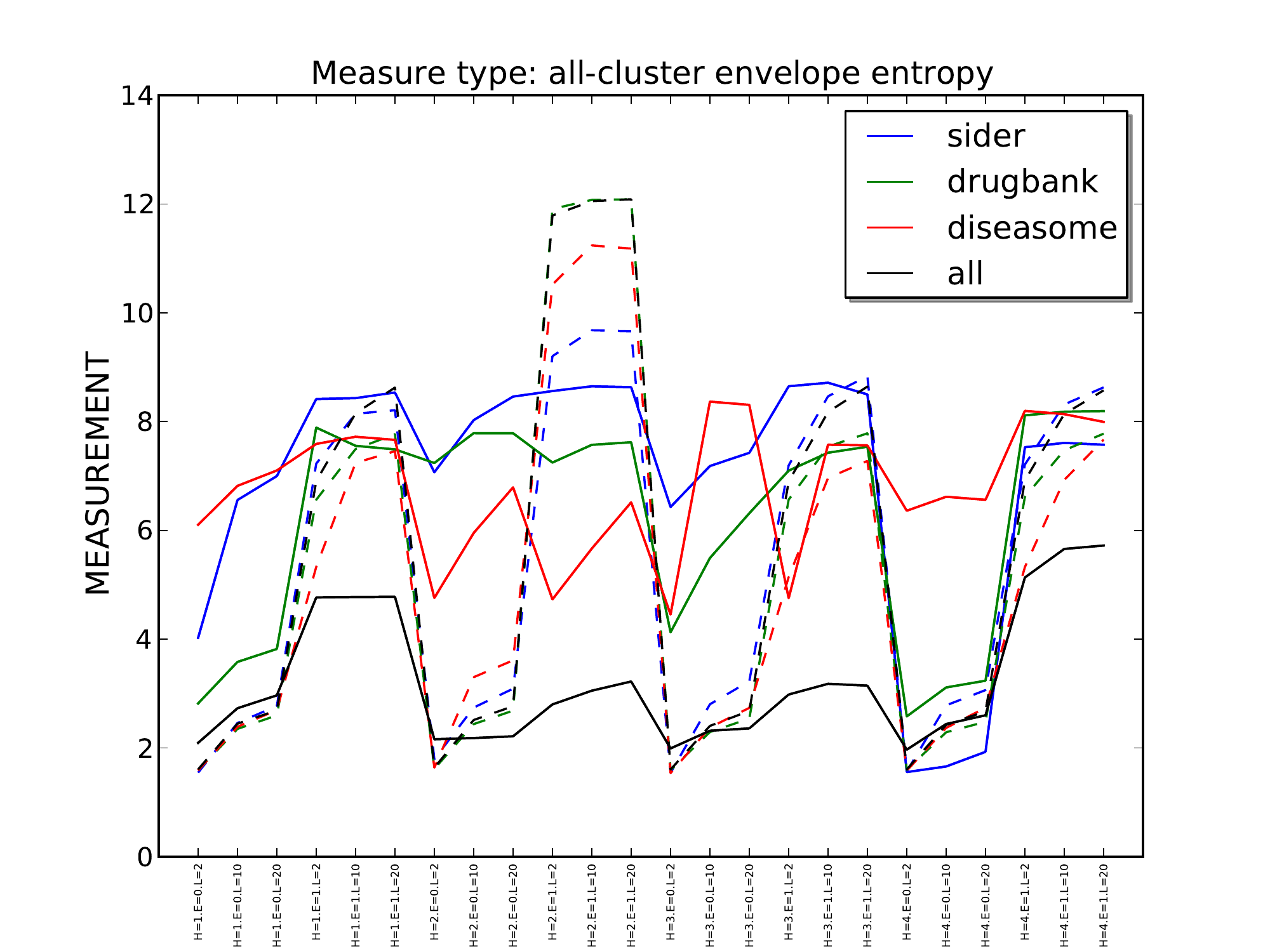}}
\caption{Entropy plots}\label{fig:entropy}
\end{figure}

The {\bf entropy} results using the $T_w, T_s$ taxonomies for the topic 
annotations are 
in Figure~\ref{fig:entropy}.
The observations can be summarised as follows:
\begin{inparaenum}[1.]
  \item The entropies computed using the $T_s$ taxonomy are always higher
  than the ones based on $T_w$ when taking into account only the most general
  identifiers of the cluster annotations (the left hand side plots). The trend
  is opposite, though not so clear, for the full (\ie specific) cluster 
  annotations. 
  \item The entropies tend to be higher for the H2, H3 heuristics (weight
  and similarity preferences). 
  \item Generally, the entropies increase with the length of the walks, 
  however, the {\it all} dataset tends to exhibit such behaviour more often
  than the others (which do so basically only in case of H2, H3 heuristics
  for top clusters).
  \item The isolated datasets tend to have higher entropies than the {\it all}
  one for specific clusters (right hand side plots), with {\it Diseasome} or
  {\it SIDER} being the most entropic ones and {\it DrugBank} usually being
  the second-highest.
  \item On the other hand, the {\it all} dataset has generally highest entropy 
  for the abstract clusters (left hand side plots) based on the $T_w$ taxonomy.
  \item The results based on the $T_s$ taxonomies are mostly close to each 
  other, however, the {\it Diseasome} and {\it SIDER} datasets tend to have 
  higher entropies than the others for the H2 and H3, H4 heuristics, 
  respectively. 
\end{inparaenum}

\vspace{-0.3cm}
\subsection{Interpreting the Results}\label{sec:experiments.interpretation}
\vspace{-0.1cm}

The 
classification
of the {\it SIDER}, {\it Diseasome}, 
{\it DrugBank} and {\it all} datasets 
is \da\da\ua, \ua\ua\da,
\da$--$, \ua\ua$-$, respectively. 
We assigned the \ua~or \da~symbols to datasets that have the corresponding 
measures distinctly 
higher or lower than at least two other datasets in more than half of all 
possible settings. We used a new $-$ symbol if there is no clearly prevalent 
higher-than or lower-than trend. 
According to the 
classification, {\it SIDER} is more 
serendipitous with simple balanced contexts and {\it Diseasome} is more focused 
around uneven sets of topics with complex structural information context. The
general classification of the {\it DrugBank} and {\it all} datasets is trickier
due to less significant trends observed. However, {\it DrugBank} is definitely
simpler (even more so than much smaller {\it Diseasome}), and the {\it all} 
dataset is more focused and complex. Another general observation is that the 
parameter settings 
typically do not influence the relative 
differences between the dataset performances. 
The only 
exceptions are $sim_{cos}$ start/end coherences and specific cluster 
entropies, but the differences do not seem to be too dramatic even there. 

The conflicting trends in coherences and entropies in case of {\it DrugBank}
and {\it all} datasets are related to the slightly different semantics of 
the particular measures within those classes. In case of coherence, the 
start/end one can be interpreted as an approximation of dataset's 
``attractiveness,'' \ie the likelihood of ending up in a similar topic no 
matter where and how one goes. The other coherences take into account 
consequent steps on the walks and thus are more related to the measure of 
average or cumulative topical ``dispersal'' across single steps. For 
entropies, the top-cluster and all-cluster entropies measure the information 
content regarding abstract and specific topics, respectively. Therefore the 
measures can exhibit different trends for datasets that have uneven 
granularity of the taxonomy levels.

To compare the results of the empirical analysis of the datasets with the 
intentions of their creators, let us start with {\it SIDER} that has been 
designed as simple-structured dataset where one can easily retrieve 
associations between drugs and their side effects. Our observations indeed 
confirm this -- {\it SIDER} is classified as relatively simple, with balanced
contexts and without any significant ``attractor'' topics. {\it Diseasome} 
focuses on capturing complex disease-gene relationships, which again 
corresponds to our analysis -- the dataset is relatively focused and complex
with rather low entropy in the contexts. Finally, {\it DrugBank} is supposed
to link drugs with comprehensive information spanning across multiple domains 
like pharmacology, chemistry or genetics, with the information usually defined 
in external vocabularies. The high start/end and low cumulative coherences 
indicate a strong attractiveness despite of frequent context switching
(\ie no matter where you start, it is likely that you will be in a drug-related
context and you will end up there again even if you switch between other topics
on the way). The low complexity measured by \acronym~indicates relatively 
simple structure of the links. This is consistent with a manually assessed
structure of DrugBank -- it contains many relations fanning out from drug 
entities while the other nodes are seldom linked to anything else than other 
drugs. 

One of the most interesting dataset-specific observations, though, is related 
to the aggregate {\it all} dataset. It is clearly most complex.
It has rather 
low start/end coherence, but generally quite high cumulative coherences. In 
addition, the abstract and specific entropies are relatively high and low,
respectively. This means that a traversing agent explores increasingly more 
distant topics, but shifting only a little at a time. The specific contextual
topics are quite unpredictable, but the abstract topics tend to be more
regular, meaning that one can learn a lot of details about few general domains
using the dataset. These characteristics make the {\it all} dataset most 
suitable for tasks like knowledge discovery and/or 
extraction of complex features associated with drugs or diseases. This is very 
useful information in the scope of our original motivations for picking the 
experimental datasets (\ie feature selection for adverse drug effect discovery 
models). 


\section{Conclusions and Future Work}\label{sec:conclusions}

We have presented \acronym, a well-founded methodology for empirical analysis
of LOD datasets. We have also described a publicly available implementation of 
the methodology. The experimental results 
demonstrated the utility of \acronym,
as it provided a meaningful automated 
assessment of biomedical datasets that is consistent with the intentions of the
dataset authors and maintainers. 

Our future work involves more scalable 
clustering 
and graph traversal algorithms that would make \acronym~readily applicable 
even to the largest LOD datasets like DBpedia or Uniprot. We also want to 
experiment with other implementations of the methodology, using and formally
analysing especially different similarities and clusterings. Another 
interesting research topic is studying correlation between the performance of 
specific SPARQL query types and particular \acronym~measure value ranges, 
which could provide valuable insights for maintainers and users of SPARQL 
end-points. We also want to work together with dataset providers in order to 
establish a more systematic and thorough mapping between \acronym~assessment 
of datasets and their suitability to particular use cases. Last but not least,
we intend to investigate other possible applications of the \acronym~measures,
such as machine-aided modelling or vocabulary debugging. 



\bibliographystyle{splncs}
\bibliography{bib/vit.bib}

\begin{thebibliography}{10}

\bibitem{harpaz2012novel}
Harpaz, R., DuMouchel, W., Shah, N.H., Madigan, D., Ryan, P., Friedman, C.:
\newblock Novel data-mining methodologies for adverse drug event discovery and
  analysis.
\newblock Clinical Pharmacology \& Therapeutics \textbf{91}(6) (2012)
  1010--1021

\bibitem{sna}
Wasserman, S., Faust, K.:
\newblock Social Network Analysis: Methods and Applications.
\newblock Cambridge University Press (1994)

\bibitem{Novacek2011iswc}
Nov\'{a}\v{c}ek, V., Handschuh, S., Decker, S.:
\newblock Getting the meaning right: A complementary distributional layer for
  the web semantics.
\newblock In: Proceedings of {ISWC'11}, Springer (2011)

\bibitem{10.1371/journal.pcbi.1000443}
Pesquita, C., Faria, D., Falcão, A.O., Lord, P., Couto, F.M.:
\newblock Semantic similarity in biomedical ontologies.
\newblock PLoS Computational Biololgy \textbf{5}(7) (2009)

\bibitem{novacek2014peerj}
Nov\'{a}\v{c}ek, V., Burns, G.A.:
\newblock {SKIMMR}: Facilitating knowledge discovery in life sciences by
  machine-aided skim reading.
\newblock PeerJ (2014) In press, see \url{https://peerj.com/preprints/352/} for
  a preprint.

\bibitem{nocetti2003connectivity}
Nocetti, F.G., Gonzalez, J.S., Stojmenovic, I.:
\newblock Connectivity based k-hop clustering in wireless networks.
\newblock Telecommunication systems \textbf{22}(1-4) (2003)  205--220

\bibitem{ahn2010link}
Ahn, Y.Y., Bagrow, J.P., Lehmann, S.:
\newblock Link communities reveal multiscale complexity in networks.
\newblock Nature \textbf{466}(7307) (2010)  761--764

\bibitem{campinas2012introducing}
Campinas, S., Perry, T.E., Ceccarelli, D., Delbru, R., Tummarello, G.:
\newblock Introducing {RDF} graph summary with application to assisted {SPARQL}
  formulation.
\newblock In: Proceedings of DEXA'12, IEEE (2012)  261--266

\bibitem{DBLP:conf/dexaw/LangeggerW09}
Langegger, A., W{\"o}{\ss}, W.:
\newblock {RDFStats} - an extensible {RDF} statistics generator and library.
\newblock In: DEXA Workshops. (2009)  79--83

\bibitem{moller2010learning}
M{\"o}ller, K., Hausenblas, M., Cyganiak, R., Handschuh, S.:
\newblock Learning from linked open data usage: Patterns \& metrics.
\newblock In: Proceedings of the WebSci'10, Web Science Trust (2010)

\bibitem{colazzo:hal-00960609}
Colazzo, D., Goasdou{\'e}, F., Manolescu, I., Roatis, A.:
\newblock {RDF Analytics: Lenses over Semantic Graphs}.
\newblock In: {Proceedings of WWW'14}, ACM (2014)

\bibitem{statistics_for_research}
Dowdy, S., Weardon, S., Chilko, D.:
\newblock Statistics for Research.
\newblock Wiley (2005)

\bibitem{ahmed2014amia}
Abdelrahman, A.T., Munoz, E., Nov\'a\v{c}ek, V., Vandenbussche, P.Y.:
\newblock Supporting knowledge discovery by linking diverse biomedical data.
\newblock In: AMIA'14 Abstracts, AMIA (2014) Submitted to AMIA'14, preprint at:
  \url{http://goo.gl/3jl8XL}.

\end{thebibliography}

\end{document}